\pdfoutput=1

\documentclass[11pt]{article}

\newif\ifcomment\commentfalse{}

\usepackage[a-1b]{pdfx}

\usepackage{framed}
\usepackage{mdwlist}
\usepackage{siunitx}
\usepackage{latexsym}
\usepackage{colortbl}
\usepackage[table]{xcolor}
\usepackage{nicefrac}
\usepackage{booktabs}
\usepackage{fnpct}
\usepackage{amsfonts}
\usepackage[T1]{fontenc}
\usepackage{bold-extra}
\usepackage{amsmath}
\usepackage{amssymb}
\usepackage{bm}
\usepackage{graphicx}
\usepackage{mathtools}
\usepackage{microtype}
\usepackage{multirow}
\usepackage{multicol}
\usepackage{placeins}
\usepackage{xpatch}
\usepackage{latexsym,comment}
\usepackage[normalem]{ulem}
\hypersetup{
    colorlinks,
    linkcolor={red!!black},
    citecolor={blue!50!black},
    urlcolor={blue!80!black}
}

\newcommand*{\missingreference}{{\colorbox{red}{?reference?}}}
\newcommand*{\missingcitation}{{\colorbox{red}{?citation?}}}

\makeatletter
\xpatchcmd{\@setref}{\bfseries}{\missingreference}{}{}
\def\@citex[#1]#2{\leavevmode
    \let\@citea\@empty
    \@cite{\@for\@citeb:=#2\do
        {\@citea\def\@citea{,\penalty\@m\ }%
            \edef\@citeb{\expandafter\@firstofone\@citeb\@empty}%
            \if@filesw\immediate\write\@auxout{\string\citation{\@citeb}}\fi
            \@ifundefined{b@\@citeb}{\hbox{\reset@font\missingcitation}%
                \G@refundefinedtrue
                \@latex@warning
                {Citation `\@citeb' on page \thepage \space undefined}}%
            {\@cite@ofmt{\csname b@\@citeb\endcsname}}}}{#1}}
\makeatother

\newcommand{\ai}[0]{\abr{ai}}
\newcommand{\mm}[0]{\abr{llm}}

\newcommand{\gem}[1]{\mbox{\textsc{gem}}}
\newcommand{\abr}[1]{\textsc{#1}}

\renewenvironment{quote}
{\list{}{\rightmargin\leftmargin}%
    \item\relax\small\ignorespaces}
{\unskip\unskip\endlist}

\newcommand{\hidetext}[1]{}
\newcommand{\ignore}[1]{}

\ifcomment
    \newcommand{\pinaforecomment}[3]{\colorbox{#1}{\parbox{.8\linewidth}{#2: #3}}}

    \newcommand{\prtodo}[1]{\pinaforecomment{lightblue}{pr}{#1}}
    \newcommand{\prtodoi}[1]{\pinaforecomment{lightblue}{pr}{#1}}
    \newcommand{\baseinlinecomment}[1]{{\textcolor[rgb]{1.0, 0.0, 0.0}{#1}}}
\else
    \newcommand{\baseinlinecomment}[1]{}
    \newcommand{\pinaforecomment}[3]{}
    \newcommand{\prtodo}[1]{}
    \newcommand{\prtodoi}[1]{}
\fi

\definecolor{lightblue}{HTML}{3cc7ea}
\definecolor{CUgold}{HTML}{CFB87C}
\definecolor{bumblebee}{HTML}{F7CE4C}
\definecolor{grey}{rgb}{0.95,0.95,0.95}
\definecolor{ceil}{rgb}{0.57, 0.63, 0.81}
\definecolor{UMDred}{HTML}{ed1c24}
\definecolor{UMDyellow}{HTML}{ffc20e}
\definecolor{darkgreen}{rgb}{0.0, 0.5, 0.0}
\definecolor{greencustom}{rgb}{0.0, 0.4, 0.0}
\definecolor{lightred}{rgb}{1.0, 0.4, 0.4}
\definecolor{lightgreen}{rgb}{0.6, 1.0, 0.6}
\definecolor{britishracinggreen}{rgb}{0.0, 0.26, 0.15}
\definecolor{lilac}{RGB}{200,162,200} 
\definecolor{pastelgreen}{RGB}{176, 201, 170} 

\newcommand{\jbgcomment}[1]{\pinaforecomment{lightred}{JBG}{#1}}
\newcommand{\mgorcomment}[1]{\pinaforecomment{pastelgreen}{mgor}{#1}}

\newcommand{\efcomment}[1]{\pinaforecomment{orange}{Eve}{#1}}

\newcommand{\mgor}[1]{\mgorcomment{#1}}

\newcommand{\smallurl}[1]{ \begin{tiny}\url{#1}\end{tiny}}

\newcommand{\llm}[0]{\abr{llm}}
\newcommand{\llms}[0]{\abr{llm}{\small s}}

\newcommand{\qa}[0]{\abr{qa}}

\newcommand{\bert}{\abr{bert}}

\newcommand{\numHumanPlayers}{23}           
\newcommand{\numAIAgents}{16}               
\newcommand{\numTournaments}{2}             
\newcommand{\numGames}{24}                  
\newcommand{\avgExperienceYears}{3.2}       

\newcommand{\numTossups}{140}               
\newcommand{\numBonusParts}{420}            
\newcommand{\numTossupsPerGame}{20}         
\newcommand{\numBonusesPerGame}{20}         

\newcommand{\numProactiveDecisions}{387}    
\newcommand{\numDeliberativeDecisions}{1440} 
\newcommand{\numMutingDecisions}{150}       
\newcommand{\numSwitchingDecisions}{450}    

\newcommand{\pointsCorrectAnswer}{10}       
\newcommand{\pointsIncorrectBuzz}{-5}       
\newcommand{\pointsCorrectBonusPart}{10}    

\newcommand{\pctAIConsensusSwitch}{82}      
\newcommand{\pctSwitchWhenWrong}{68}        
\newcommand{\pctExplanationImprovement}{12} 

\newcommand{\pctEarlyRoundError}{28}        
\newcommand{\pctLateRoundError}{18}         
\newcommand{\pctCalibrationImprovement}{10} 

\newcommand{\betaConfidence}{1.85}          

\newcommand{\writerCompensation}{25}        

\newcommand{\pctOracleAchieved}{79}         
\newcommand{\pctTopicMutingIncrease}{30}   
\newcommand{\pctEarlyMutingTiming}{49}      
\newcommand{\pctEarlyMutingDecisions}{18}    
\newcommand{\pctOptimalMutingDecisions}{9}    
\newcommand{\pctLateMutingDecisions}{73}    
\newcommand{\mutingTeamLost}{T2}            

\newcommand{\pctHumanBuzzedFirst}{17.9}
\newcommand{\pctHumanInterruptIncorrect}{20.0}
\newcommand{\pctAIInterruptIncorrect}{29.4}
\newcommand{\bonusAccHuman}{42.8}
\newcommand{\bonusAccAIRandom}{59.4}
\newcommand{\bonusAccAIOracle}{77.6}
\newcommand{\bonusAccBestPick}{81.2}
\newcommand{\bonusAccFinal}{81.7}
\newcommand{\pctUnderReliance}{3.9}

\newcommand{\pctOverReliance}{1.7}

\newcommand{\pctUnderRelianceHaiC}{64.5}

\newcommand{\statSynergyMcNemar}{671.68}
\newcommand{\statSynergyP}{<0.001}
\newcommand{\statConsensusChi}{9.95}
\newcommand{\statConsensusP}{0.002}
\newcommand{\statConfirmBiasChi}{Fisher}
\newcommand{\statConfirmBiasP}{<0.001}
\newcommand{\statTemporalBeta}{0.41}
\newcommand{\statTemporalP}{0.01}
\newcommand{\statDiscernEarly}{27.1}
\newcommand{\statDiscernLate}{75.0}

\usepackage{style/acl}
\usepackage{lmodern}

\usepackage{times}
\usepackage{bookmark}
\usepackage[T1]{fontenc}

\usepackage[utf8]{inputenc}

\usepackage{microtype}
\usepackage[most]{tcolorbox} 
\newtcolorbox{designprinciple}[2]{
  enhanced,
  left=4pt, right=4pt, top=3pt, bottom=3pt,
  colback=teal!6,
  colframe=teal!40!black,
  boxrule=0.4pt,
  borderline west={2.5pt}{0pt}{teal!50!black},
  arc=3pt,
  fonttitle=\bfseries\small,
  colbacktitle=teal!15,
  coltitle=teal!70!black,
  title={\abr{dp #1}: #2},
  before skip=6pt, after skip=6pt,
}
\usepackage{array}
\usepackage{tabularx}
\usepackage{lipsum} 

\newcommand{\systemUrl}[1]{https://github.com/qanta-org/qb-tournament-runner}
\newcommand{\datasetUrl}[1]{https://huggingface.co/datasets/qanta-challenge/qanta25-gamedata}
\newcommand{\githubUrl}[1]{https://github.com/qanta-org/qanta25-analysis}

\newcommand{\guess}[1]{\uline{#1}}
\newcommand{\answer}[1]{\textbf{\uline{#1}}}

\newcommand{\RodeRunner}{System~1}
\newcommand{\BlackRaven}{System~2}
\newcommand{\Tigerclaw}{System~3}
\newcommand{\WiseWings}{System~4}
\newcommand{\Sphinx}{System~5}
\newcommand{\Sam}{System~6}
\newcommand{\Magicarp}{System~7}
\newcommand{\Bayleef}{System~8}

\newcommand{\teaminpersfour}{Team~4}

\newcommand{\teamonlinefour}{Team~8}
\newcommand{\teamonlinefive}{Team~9}

\newcommand{\gpt}[1]{\abr{gpt}-{\small#1}} 

\newcommand{\draftSelection}[0]{team formation}

\newcommand{\tossup}[0]{tossup}
\newcommand{\Tossup}[0]{Tossup}

\newcommand{\bonus}[0]{bonus}
\newcommand{\Bonus}[0]{Bonus}

\definecolor{qbred}{rgb}{0.5, 0.0, 0.0}
\definecolor{agentblue}{HTML}{3366CC}
\definecolor{agentgreen}{HTML}{109618}
\definecolor{agentpurple}{HTML}{990099}
\definecolor{agentlightblue}{HTML}{0099C6}
\definecolor{agentpink}{HTML}{DD4477}
\definecolor{agentdarkblue}{HTML}{316395}
\definecolor{caimirayellow}{HTML}{E69F00}
\definecolor{caimiradarkyellow}{HTML}{CEA146}
\definecolor{caimiragreen}{HTML}{509364}
\definecolor{caimiradarkgreen}{HTML}{3c6c4c}
\definecolor{dullred}{HTML}{953131}
\definecolor{dullblue}{HTML}{175B98}

\title{AI, Take the Wheel: What Drives Delegation and Trust in Human--Computer Cooperative Question Answering?}

\author{
Maharshi Gor, Yoo Yeon Sung, and Yu Hou \\
\mdseries University of Maryland \\
\mdseries College Park, MD, USA
\And
Eve Fleisig \\
\mdseries University of California \\
\mdseries Berkeley, CA, USA
\AND
Irene Ying \\
\mdseries Phasechange.ai \\
\mdseries Lakewood, CO, USA
\And
Tianyi Zhou \\
\mdseries MBZUAI \\
\mdseries Abu Dhabi, UAE
\And
Jordan Boyd-Graber \\
\mdseries University of Maryland \\
\mdseries College Park, MD, USA
}

\begin{document}
\maketitle

\jbgcomment{prefer to have e-mails than town}

\begin{abstract}
AI systems are fallible, and humans can make mistakes in deciding whether to trust
AI over their own judgment.
Thus, improving human-AI collaboration requires understanding \emph{when},
\emph{why}, and \emph{how} humans decide to rely on AI.
We study two distinct reliance decisions: the delegation choice---deciding
when to let \ai{} act autonomously without knowing its output, and
the adoption choice---evaluating \ai{} suggestions and deciding how to use them.
Both of these decoupled reliance patterns shape collaboration, but prior
work rarely studies them together in realistic settings with the same users.
We address this gap by studying collaborative human--\ai{} teams competing in a
question-answering game in which humans can choose when and how to work with \ai{} agents to win.
Our \numGames{} matches pair \numHumanPlayers{} expert
humans with \numAIAgents{} \abr{ai} agents, capturing
\numProactiveDecisions{} delegation and
\numDeliberativeDecisions{} adoption decisions.
While human-\ai{} collaboration performs better than either \ai{} or humans
alone, humans make suboptimal collaboration decisions, both
under-relying on correct \ai{} suggestions
(\pctUnderReliance{}\% of opportunities missed) and
over-relying when \abr{ai} misleads them
(\pctOverReliance{}\%).
Both parties contribute wrong answers: reported model confidence is near chance when humans and \ai{} disagree, while confirmation bias drives higher
under-reliance (\pctUnderRelianceHaiC{}\%) when an
\ai{} suggestion agrees with humans' initial incorrect answer.
To close this gap, we recommend calibrated confidence,
evidence-grounded explanations, and mechanisms that help
users refine trust.
\end{abstract}
\section{Introduction: How much do you trust \mm{} output?}

A wide range of people---from casual users looking for basic explanations to domain experts requiring professional support in
medicine~\cite{Leonard_2024}, law~\cite{magesh2024ai}, and
finance~\cite{maple2024impact}---are using \ai{}.\footnote{Throughout
  the paper, we use \ai{} as a stand-in for text-based agentic workflows
  that are built using transformer-based \llms{} (e.g., \gpt{4.1}, Claude 3.5, etc.); details in Appendix~\ref{appendix:collabqa-ai-systems}.}
%
However, some users  over-trust \ai{} answers because of limited time or imperfect knowledge of
\ai{} capabilities~\cite{goddard2012automation,bansal2019beyond,bucinca2021trust},
%
while other users are skeptical of \ai{}
output when it would indeed help them~\cite{kleinberg2018human,jakesch2023cowriting}.

\begin{figure}[t]
    \centering
    \includegraphics[width=\columnwidth]{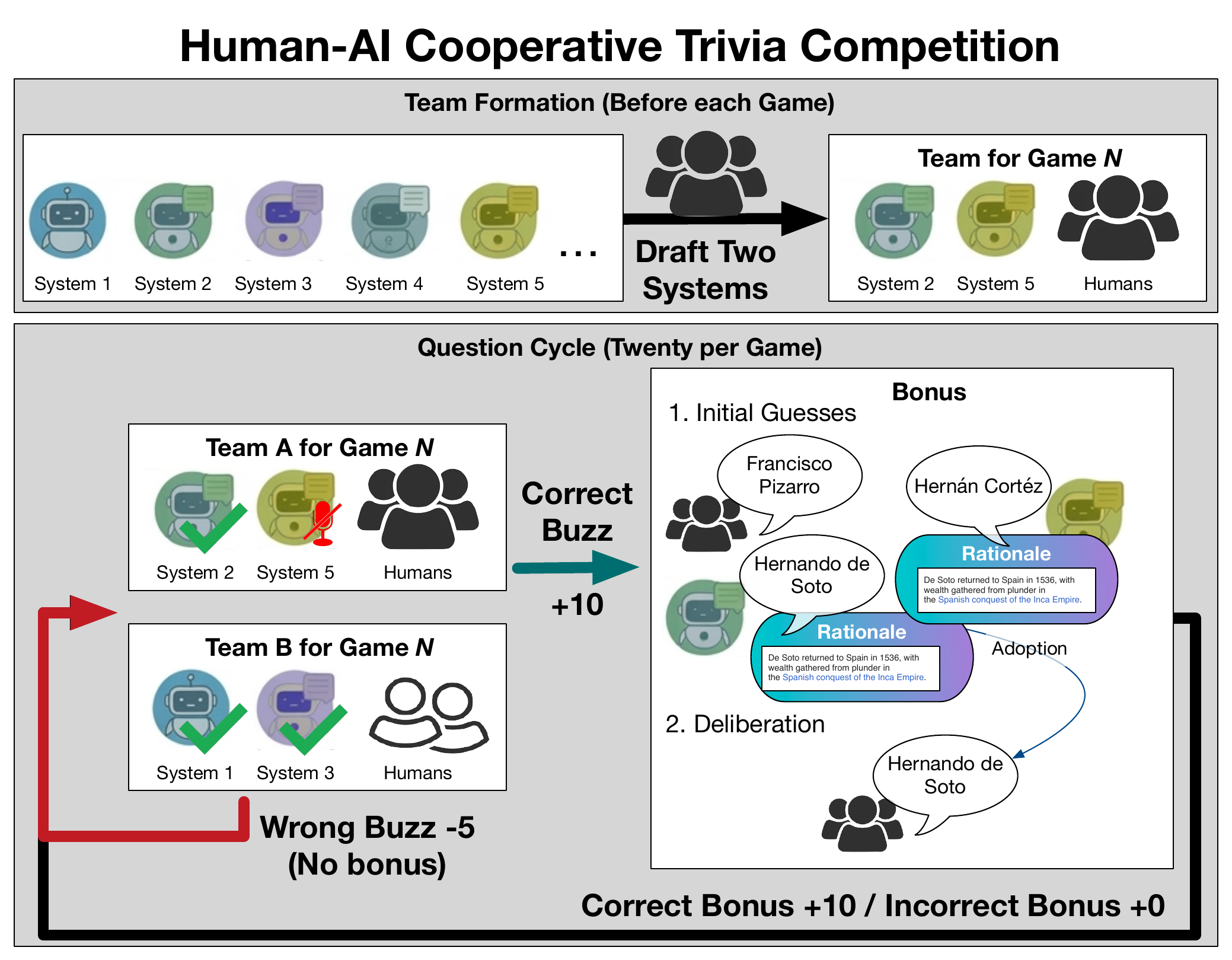}
    \caption{Our experimental setup has humans working with \ai{}
      teammates in two competitive \qa{} settings. In the \tossup{}
      phase, \ai{} teammates can directly answer questions without
      human intervention. In the \bonus{} phase, humans and \ai{}
      collaborate to reach consensus, with humans providing the final
      answer.}
    \label{fig:collabqa-teaser}
\end{figure}

Prior work~\cite{lee2004trust} focuses on trust and adoption of \mm{}
output through post-hoc acceptance measures or synthetic tasks
that overlook real-world reliance dynamics and strategic choices
under uncertainty (Section~\ref{sec:collabqa-related}).
A recent taxonomy of 66 human--\ai{} studies
confirms that capturing both delegation and advisory patterns in
a single study remains rare~\cite{gomez2025humanai}.
Moreover, real-world reliance decisions happen under pressure: limited time,
imperfect knowledge of \ai{} capabilities (even among experienced users),
existing human social dynamics, and
limited chances to learn through repeated interaction.

We address this gap by studying how skilled users in human-\abr{ai}
\emph{teams} use \ai{} outputs in a live \textbf{novel collaborative
  benchmark}\footnote{Dataset:~\small{\url{\datasetUrl{}}};\\
  Platform:~\small{\url{\systemUrl{}}};\\
  Analysis:~\small{\url{\githubUrl{}}}.}
that measures \textit{when} and \textit{how} humans
rely on \ai{} assistance through two distinct reliance decisions.
\emph{Proactive delegation} is the decision to let \ai{} act
autonomously without reviewing its output---capturing \textit{when}
\abr{ai} will be used.
\emph{Deliberative adoption} is the decision to accept or reject
\ai{} output after evaluating it---capturing \textit{how} \abr{ai}
output shapes final decisions.

We examine these reliance decisions through a competitive trivia
tournament~(Section~\ref{sec:collabqa-quizbowl}). First, humans form teams with
\ai{} teammates~(Figure~\ref{fig:collabqa-teaser}).
In \textbf{team formation}, teams draft \ai{} systems from a pool
of based on perceived capabilities, reflecting humans' perception of \ai{} capabilities. 
There are two types of questions: \tossup{}s and \bonus{}es, that captures the two
distinct reliance decisions.
In the \textbf{\tossup{}} phase (proactive delegation), teams decide
whether to let their chosen \ai{} system answer autonomously or
mute it entirely, revealing beliefs about \ai{} reliability without oversight.
In the \textbf{\bonus{}} phase (deliberative adoption), teams see \ai{}
suggestions with confidence scores and explanations, then decide
whether to adopt them, showing what model outputs contribute to trust and how useful those outputs are.
These three scenarios capture complementary facets of human--\ai{}
collaboration.
Analyzing all three provides insights into ``how''
people rely on \ai{}: not just whether they accept or reject
suggestions, but which models they trust, when, and why.
%
%
%
%
For collaboration to work, the questions must require both human and computer skills~(Section~\ref{sec:collabqa-question-design}).

While collaboration is mostly
synergistic---better than either alone---it is not perfect
(Section~\ref{sec:collabqa-analysis}).
\textbf{Teams miscalibrate trust}, primarily through
under-reliance: they fail to adopt correct \ai{} answers
\pctUnderReliance{}\% of the time when they initially propose incorrect answers.
Over-reliance is less common (\pctOverReliance{}\%), but teams
override their correct answers with incorrect \ai{} suggestions.
Both parties contribute to these errors. On the \textbf{\ai{} side}, confidence
scores are poorly calibrated: when humans and \ai{} disagree, relying on model
confidence to select the correct answer performs near chance.
%
On the \textbf{human side}, confirmation bias amplifies mistakes---when an
incorrect human answer is confirmed by one of the two \ai{} teammates,
under-reliance rises to \pctUnderRelianceHaiC{}\% as agreement signals reinforce wrong judgments.
High-skill teams are particularly susceptible to this effect, as expertise
breeds overconfidence in initial judgments.

The successful collaborations provide \textbf{a blueprint for
improving future human--\ai{} collaboration}: we distill five
design principles throughout our analysis
(Section~\ref{sec:collabqa-analysis}).
%
In the \bonus{} phase, explanations that cite specific question clues help
humans abandon wrong answers \pctExplanationImprovement{}\% more often.
However, we note that features that predict \ai{} correctness
(reasoning coherence, question understanding) differ from what
humans trust (surface similarity, presence of quotes).
Encouragingly, human teams improve with practice: across \bonus{} rounds, inaccuracies decrease
from 28\% to 18\%.
Most strikingly, teams reach correct answers in 5.5\% of
cases where neither humans nor \ai{} were initially right.
%
%


\section{Game Design for Human--AI Collaborative Question Answering}
\label{sec:collabqa-quizbowl}

We study human-\ai{} collaboration through a competitive trivia tournament where
mixed teams face off in question-answering games.
Each team has up to three human players and two \ai{}s.
Two teams compete head-to-head in a game that alternates between two question
types: ``\tossup{}'' questions, where any player from either team can buzz in to
answer individually, and ``\bonus{}'' questions, where the team that answered the
\tossup{} correctly collaborates on a three-part question.
Before gameplay, teams draft which \ai{} agents to work with from a pool of
available systems (Section~\ref{sec:collabqa-tourney-structure}).

This design is grounded in the trivia domain~\cite{joshi-17}, which provides
both challenging questions and a motivated, enthusiastic
community~\cite{jennings-06}.
\citet{koivisto2019rise} argue that empirical human data collection is more
effective when participants are intrinsically motivated.
Unlike synthetic
laboratory studies, competitive trivia tournaments provide players who are familiar with the task, face real stakes with
costs for incorrect reliance on agents, and deliberate with other humans.
Our format is based on Quizbowl~\cite{boyd2012besting}, a well-established
academic trivia competition.

\noindent\textbf{Game flow.}
A game is played between two teams and consists of \numTossupsPerGame{} cycles.
Each cycle begins with a \tossup{} question: any player can buzz in to answer, and
a correct answer earns \pointsCorrectAnswer{} points for the team (an incorrect
buzz before the question ends costs \pointsIncorrectBuzz{} points).
If a team answers the tossup correctly, they earn a \bonus{} question with three
parts, each worth \pointsCorrectBonusPart{} points.
The team collaborates to answer each part, with humans making the final
decision after seeing \ai{} suggestions.
The rest of this section details each phase and the human--\ai{}
collaboration.

\subsection{\Tossup{} Delegation (Tossups)}
\label{sec:collabqa-tossup-setup}


%
%
%

%
\Tossup{}s are designed to be \emph{interrupted} by a \textbf{buzz}.
Each \tossup{} question (a sequence of clues starting hard and getting easier) is read aloud to all players.\footnote{This decrease in difficulty should be true for both types of players (humans and \ai{}); see Section~\ref{sec:collabqa-question-design} for how we got trivia experts to write these questions.}
Once any player---human or
\ai{}---is confident enough to answer, they ``buzz'' in with a
response. 
Humans buzz using a physical buzzer (or its online
equivalent) and must answer immediately with no team discussion,
following standard quizbowl rules. An \ai{} buzzes in using the interface and similarly immediately vocalizes an answer.
However, each team only has one chance to answer a tossup; an
incorrect guess results in a penalty and bars the incorrect
player's whole team from answering the question and participating
in the \bonus{} phase.
Thus, a player with low accuracy or poor calibration hinders the whole team.
Human teammates can shout or glare at poorly calibrated teammates;
for \ai{} teammates, we provide an analogous mechanism: \textbf{muting}.

\paragraph{Muting \ai{} teammates.}
Teams may \textbf{mute} an \ai{} teammate at the start of the game or
after any \tossup{}--\bonus{} cycle,\footnote{A cycle ends immediately
  before the next tossup question is read. If neither team answered
  the tossup correctly, the cycle ends right after the tossup.}
preventing it from buzzing for the rest of the game.
Crucially, humans still see muted \ai{}s' suggestions in the \bonus{}
phase---teams may distrust an \ai{}'s buzzing judgment while still
valuing its knowledge for collaborative decisions.
For instance, during one tournament, a question began as follows:
\begin{quote}
  In one work, a large-headed man with a multi-colored top hat and pink\dots
\end{quote}
One of the \ai{} players buzzed at this point with the guess \guess{Willy Wonka}\footnote{We distinguish a
\guess{guess}, a (possibly incorrect) response, from
the correct \answer{answer} to the question.} with 90\%
confidence. The correct answer was \answer{Christ}.
This egregious mistake led to an immediate request to mute that \ai{} teammate for the
rest of the game.
%
%

%


\subsection{\Bonus{} Adoption (Bonuses)}
\label{sec:collabqa-bonus-setup}
\begin{figure}[t]
    \centering
    \includegraphics[width=\columnwidth]{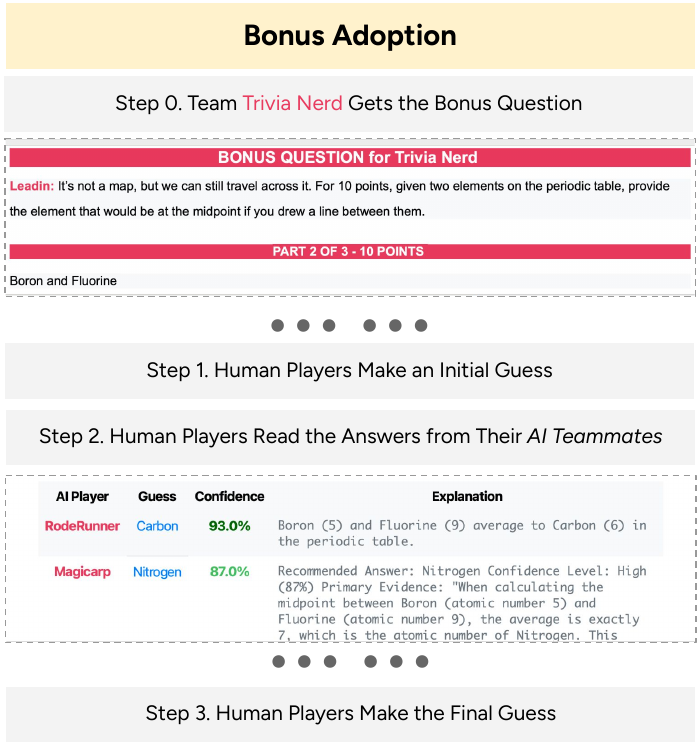}
    \caption{Overview of our collaboration interface showing deliberative 
    decision-making for bonus questions (top). Humans first provide their own guess without
    any assistance from \ai{}, then see suggestions from two \ai{} teammates
    with confidence scores and explanations (middle),
    and finally discuss and decide how to adopt them (bottom).}
    \label{fig:bonus-interface}
\end{figure}

When a team answers a tossup correctly, they earn a ``bonus'' question with three parts
on a common theme~\cite{Elgohary-18} announced with a ``lead-in'' (e.g.,
Figure~\ref{fig:bonus-interface}, top).
%
To measure the effect of collaboration, we want to measure human
ability in the absence of \ai{} assistance \emph{and} how they navigate
(often unreliable) \ai{} support.
This two-stage design is essential: by recording the human guess
\emph{before} revealing \ai{} suggestions, we can directly compare
the same team's pre- and post-\ai{} answers, isolating the causal
effect of the \ai{} on human decisions.
%

\noindent\textbf{Initial Human Guess.} 
The moderator reads each part of the question, allows the human
players to confer on the answer, and the teams provide a consensus
guess without any assistance from their \ai{} teammates.
In the illustrated example~(Figure~\ref{fig:bonus-interface}), the second part of the question
asks the team to find the element that is the midpoint of Boron and Fluorine on the
periodic table.
Human players gave a correct guess of \guess{nitrogen}.
While we record this guess and its correctness for analysis, this does
not contribute to the team's score.
The human team is not told whether their guess is correct or not.

\noindent\textbf{\ai{} Guesses.}
%
%
After the humans guess, they then see guesses from two distinct
\ai{} systems they drafted to be their teammates, along with a
confidence score and textual
explanation~(Figure~\ref{fig:bonus-interface}, middle). These
explanations are generated by the \ai{} agents themselves---not
designed by the experimenters---creating natural heterogeneity
across the \numAIAgents{} systems built by different participants
using diverse architectures.
Here, \RodeRunner{} the \ai{} nicknamed RodeRunner\footnote{Teams know their \ai{} teammates only by opaque
  nicknames during drafting (Section~\ref{sec:collabqa-tourney-structure})}
uses correct reasoning, but incorrect math, leading to an incorrect
guess of \guess{carbon} with a high confidence score of 93\%, while
the other system (Magicarp) guesses correctly \guess{nitrogen} with
a clear explanation and a confidence score of 87\%.

\noindent\textbf{Final Consensus Guess.}
The humans on the team now need to give a final answer.
They can: retain their initial guess, pick one of the  \ai{} guess(es),
or make a new guess entirely (after deliberation).
This final answer counts for the team's score.
In this case, the team kept their \guess{nitrogen} guess, confirmed by
Magicarp, earning 10 points.

\subsection{Behavioral Traces Captured}
\label{sec:collabqa-reliance-signals}

The game design yields three complementary behavioral traces.
For \textbf{team formation}, we record each team's agent selections
and the draft pick order for every round
(\S~\ref{sec:collabqa-tourney-structure}).
For \textbf{tossups} (proactive delegation), we record who buzzed
(human or \ai{}), the clue position, each \ai{}'s mute state,
buzz correctness, and points gained or lost.
For \textbf{bonuses} (deliberative adoption), we record the human
team's initial answer, each \ai{}'s answer together with its confidence score (0--1)
and textual explanation, the team's final answer, and correctness at each stage.
Together these traces span the full arc of reliance decisions, from
team formation through proactive delegation to deliberative adoption.
\section{Human--\ai{} Cooperative Trivia Tournament}
\label{sec:collabqa-tournament}


Following the game design (Section~\ref{sec:collabqa-quizbowl}), we ran two
tournaments (one in-person, one online) with \numHumanPlayers{} human players
and \numAIAgents{} \ai{} agents.
This section outlines tournament structure~(\S~\ref{sec:collabqa-tourney-structure}),
question design~(\S~\ref{sec:collabqa-question-design}), and \ai{}
agent architectures~(\S~\ref{sec:collabqa-ai-agents}).

\subsection{Tournament Structure and Participants}
\label{sec:collabqa-tourney-structure}

The tournament ranks human trivia teams on both their knowledge and
their ability to form and work with \ai{} teams.
Both tournaments---in-person and online---collect behavioral traces
of human--\ai{} collaboration.\footnote{We compared online and
in-person teams on bonus accuracy, switching rate, and muting
behavior and found no significant differences, so we pool them
in all analyses.}
Our participants comprise \numHumanPlayers{} experienced trivia
players whose competitive experience ranges from 1 to 7+ years
(mean \avgExperienceYears{}), including several with national
game show appearances.
They formed nine teams across both tournaments.

\noindent\textbf{Team formation.}
The tournament is organized into games, each featuring a themed
question packet (e.g., music, spatial reasoning, cultural references).
Before each round, teams select two \ai{} agents as teammates using a
serpentine draft~\cite{lee2022drafting}:
pick order ascends from the lowest- to the highest-scoring team,
which picks twice consecutively before the order reverses back down
to the lowest-scoring team.
Each \ai{} agent can only be selected once per round, preventing all teams from
choosing the perceived best \ai{} and avoiding conflicts when identical agents
would face each other.\footnote{Draft mechanics varied slightly between
  prelim phase and playoff rounds to give teams access to preferred \ai{} teammates; see
  Appendix~\ref{appendix:collabqa-draft-mechanics} for details.}
%
This design gives weaker teams first access to the perceived-best \ai{}
teammates, partially offsetting human skill gaps and preventing
runaway advantages by stronger teams.

Human players initially know nothing about their potential \ai{}
teammates. The models' origins are obscured through opaque
nicknames (e.g., ``RodeRunner,'' ``Magicarp''). Over the course of
early rounds, teams observe which \ai{}s buzzed and whether they were correct;
when they win a tossup, they additionally see their own \ai{} teammates'
answers, confidence scores, and explanations for that \bonus{}.
Teams use this accumulating behavioral evidence to inform drafting choices in
subsequent rounds.
Teams must decide not only \textit{whether} to trust \ai{}, but
\textit{which} \ai{} to work with---a more realistic reflection of
real-world AI adoption than studies with single fixed systems.

\noindent\textbf{Game Structure and Schedule.}
Each game consists of \numTossupsPerGame{} \tossup{} questions and
\numBonusesPerGame{} three-part \bonus{} questions.
Teams play \numGames{} games across the two tournaments, yielding
\numTossups{} toss-ups and \numBonusParts{} bonus parts.
The format includes a round-robin phase followed by single-elimination
playoffs; each game uses a themed packet targeting known \ai{}
or human weaknesses (temporal reasoning, cultural references,
wordplay, etc.).
Appendix~\ref{appendix:collabqa-dataset-overview} summarizes the
dataset scale.
\vspace{-0.5em}
\subsection{Adversarial Question Design}
\label{sec:collabqa-question-design}

\begin{figure}[t]
    \centering
    \includegraphics[width=\columnwidth]{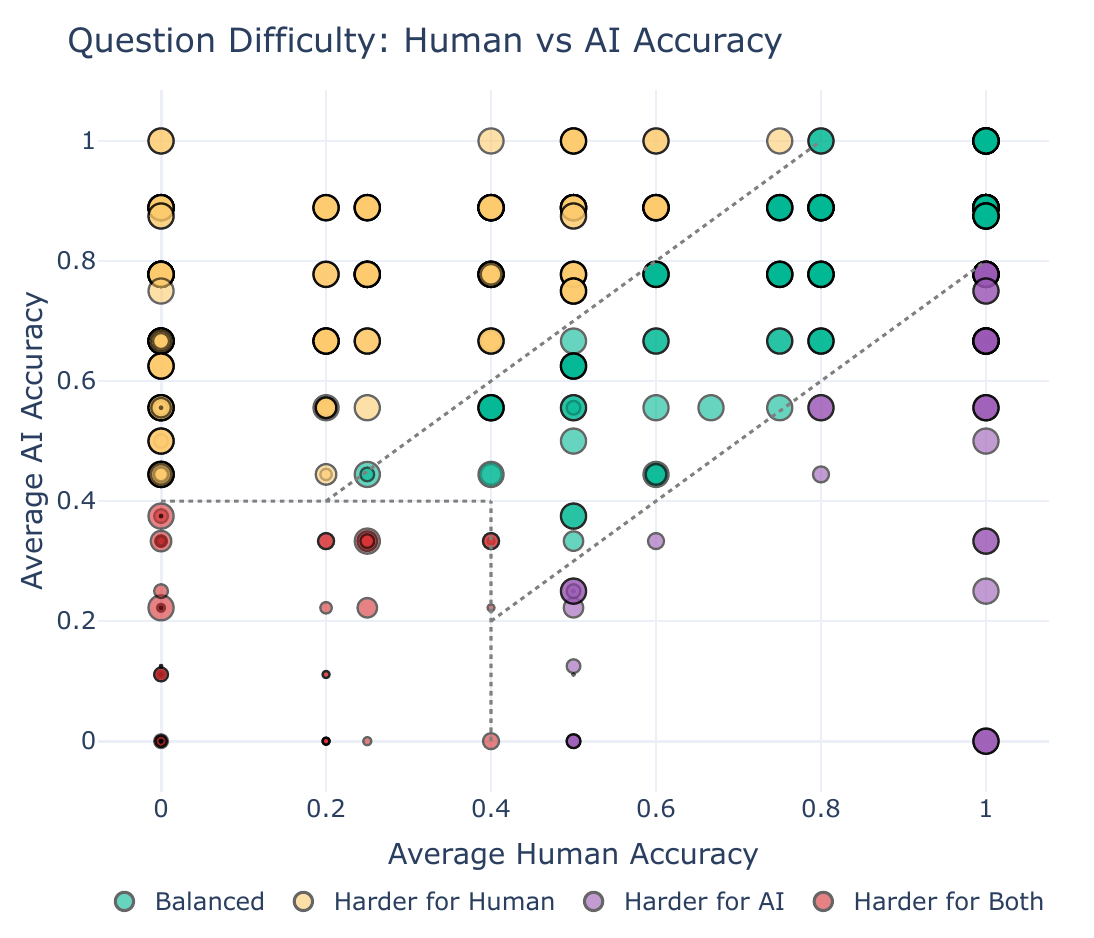}
    \caption{Question difficulty reveals systematic complementarity
    between humans and \ai{}. Each point is a question; x-axis shows
    average human accuracy, y-axis \ai{} accuracy. Bubble size
    indicates team accuracy after AI-assisted deliberation.
    Collaboration generally improves accuracy, but opportunities
    remain for questions that challenge both parties.}
    \label{fig:collabqa-question-difficulty}
\end{figure}

We adopt the adversarial question-writing framework of
\citet{sung-etal-2025-grace}, which uses human-in-the-loop
authoring~\cite{kiela2021dynabench, Wallace-19b} to create questions
challenging for both humans and \ai{} systems.
\Tossup{} questions must be difficult \emph{at every clue} while
still decreasing in difficulty, since they can be
interrupted~(Section~\ref{sec:collabqa-tossup-setup}); \bonus{}
questions must be hard enough that neither party can answer
trivially alone, yet designed so each side's strengths compensate
for the other's blind spots.
Humans struggle with precise factual recall or
cross-domain linking, while \ai{} systems falter on culturally
embedded reasoning and indirect
references~\citep{gor-etal-2024-great}, making solo success
unlikely for either party.
Figure~\ref{fig:collabqa-question-difficulty} shows bonus question
difficulty for humans and \ai{}s, highlighting this systematic
complementarity.
Appendix~\ref{appendix:collabqa-adversarial-questions} lists round
themes and examples.

\subsection{AI Agent Architectures}
\label{sec:collabqa-ai-agents}

The tournaments feature \numAIAgents{} distinct \ai{} agents built
through a four-week open competition before tournament play.
Architectures range from single-model calls with engineered prompts
to multistep pipelines with up to four model consultations for
answer generation, verification, and confidence calibration.
Base models include GPT-4.1, GPT-4o, Claude~3.5 Sonnet, DeepSeek~V3,
and Cohere Command-R, often combined within a single agent.
This architectural diversity produces heterogeneity in capabilities:
agents range from 30 to 80\% accuracy on our question set, with markedly
different strengths across question types, domains, and confidence
calibration, ensuring no single agent dominates and teams lack an obvious drafting strategy (Section~\ref{sec:collabqa-tourney-structure}).
Full agent specifications appear in
Appendix~\ref{appendix:collabqa-ai-systems}.


\section{How Humans Trust \ai{} Assistance---and Where They Misjudge}
\label{sec:collabqa-analysis}

We analyze how teams navigate the two forms of reliance
our tournaments capture.
Section~\ref{subsec:collabqa-tossup-analysis} examines proactive
delegation via muting decisions on tossups: what drives teams
to mute their \ai{}, whether muting helps, and how close teams
come to optimal muting.
Section~\ref{subsec:collabqa-bonus-analysis} examines deliberative
adoption on bonuses: how teams evaluate \ai{} suggestions, what
drives switching, and where calibration breaks down.
We distill findings into design principles (\abr{dp})
for human--\ai{} collaborative systems, highlighted in
colored callouts.

The trivia setting kept users engaged: they answer questions
before the \ai{}s, 
\jbgcomment{How many?\\\mgor{We later reiterate giving numbers in Section 4.1}}
offered guesses on their own, and
strategically muted \ai{} teammates when beneficial.
Muting rates range from 30\% to 100\% depending on round
theme, and answer adoption on bonuses ranged from 28\% to 86\%---both
suggesting deliberate, context-sensitive decisions rather
than random behavior or blind acceptance.
Despite being mostly synergistic, collaboration is not
frictionless. Under-reliance (\pctUnderReliance{}\% of
help opportunities missed) exceeds over-reliance
(\pctOverReliance{}\%), with confirmation bias and
cross-model calibration failure as primary drivers.
Key comparisons use chi-squared and McNemar's tests with
effect sizes reported where appropriate.

\subsection{\Tossup{} Delegation: Muting Decisions}
\label{subsec:collabqa-tossup-analysis}

Skilled humans often buzzed before \ai{} players: \pctHumanBuzzedFirst{}\%
of tossup questions were answered by a human before any \ai{} buzz.
Humans also showed superior calibration, with only
\pctHumanInterruptIncorrect{}\% of their buzzes being incorrect
compared to \pctAIInterruptIncorrect{}\% for \ai{s}.

%
%
Beyond individual buzzes, the more interesting aspect of human--\ai{}
\emph{collaboration} is whether teams \emph{allowed} \ai{} teammates to
answer autonomously.
As described in Section~\ref{sec:collabqa-tossup-setup}, players can ``mute'' an
\ai{} teammate before a question is read, preventing it from buzzing during rest of the round.
We evaluate muting effectiveness via counterfactual
estimation: for each muting decision, we estimate what would
have happened had the \ai{} remained active (or been muted
earlier), comparing actual outcomes against this
counterfactual and against an oracle with perfect
foresight~(Section~\ref{subsec:collabqa-tossup-analysis}).
%



\noindent\textbf{What Drives Muting Decisions?}
Teams mute less when models perform well,
with \textit{recent accuracy} being the strongest predictor.
%
%
Teams also adapt to \ai{} weaknesses, muting up to \pctTopicMutingIncrease{}\%
more on challenging topics like music, spatial reasoning
and cross-domain identification, where \ai{}s falter more.
%

\noindent\textbf{Does Muting Help?}
We analyzed whether users' mental models of \ai{} allowed them to mute optimally.
In a round of twenty questions, there are twenty opportunities to mute.
%
%
For each opportunity, we estimate the counterfactual effect on total points in the
\tossup{} phase if the \ai{} had been muted or not.\footnote{This counterfactual has
  limitations: when \ai{} buzzes early and wrong, we cannot know if a human
  would have answered correctly. However, removing an early wrong response
  is a net positive. Cases where \ai{} could have been early and right are
  straightforward to evaluate.}
%
We define an \emph{oracle policy} as the muting strategy that maximizes net
\tossup{} points\footnote{For the rest of this section, we discuss only
  \emph{net} points that muting changes, ignoring effects on subsequent
  \bonus{} questions.
  \jbgcomment{Use this footnote to give the maximum range of effects so
    that what follows has a scale.\\\mgor{I changed the points scale to percentages of optimal to avoid having to explain this.}}}
given perfect knowledge of questions and \ai{} behavior---a ceiling that
real teams cannot reach since they lack advance knowledge.
Muting pays off: eight out of nine teams earned
more points per game on average by muting than they would have by leaving
their \ai{} companions active throughout.
These teams captured \pctOracleAchieved{}\% of the oracle's maximum possible gain.
The exception, team~\mutingTeamLost{}, a strong human team, muted too 
aggressively and missed opportunities to benefit from \ai{} buzzes.

\noindent\textbf{Calibration Gaps.}
Muting is not just about silencing weak models---it requires
\textit{calibrating trust}.
%
Humans learn but remain imperfect.
Only \pctOptimalMutingDecisions{}\% of muting decisions are made
at the optimal time. Teams tend to mute late:
\pctLateMutingDecisions{}\% of muting decisions occur later than
the oracle policy would recommend. By that
point, the \ai{} may have already cost the team points through
incorrect buzzes.
In contrast, when teams do mute early (\pctEarlyMutingDecisions{}\%
of decisions), they do so with high magnitude:
\pctEarlyMutingTiming{}\% earlier in the round than optimal
(9.8 questions earlier on average).
This asymmetry is notable: \textbf{over-reliance is more frequent
but smaller in magnitude, while under-reliance is rarer but more
severe.}
As net effect, the average muting occurs 3.4 questions earlier than optimal (about
15\% of the round), suggesting early accuracy drops prompt premature
\ai{} abandonment.
The rarity of optimal timing highlights how hard it is for
humans to calibrate trust in real time, even with direct
behavioral feedback. We did note that on topics where \ai{} excels---literature, military
history---teams appropriately let \ai{} answer.
%

\begin{designprinciple}{1}{Give users granular control over
\ai{} involvement.}
\small Provide context-dependent toggles (by topic,
difficulty) rather than binary on/off controls. User agency
over \emph{when} \ai{} participates matters as much as
whether to follow its advice~\cite{vaccaro2018illusion}.
\end{designprinciple}

\subsection{\Bonus{} Answer Adoption}
\label{subsec:collabqa-bonus-analysis}

In the \bonus{} phase, teams answer multipart questions with the help of \ai{}: 
humans provide an initial guess; see \ai{} guesses,
confidence scores, and explanations; and then decide
the \emph{team's} final guess.
%

But, first: how do teams decide whether to trust \ai{} answers,
and how well do they take \ai{} advice?
To aid our analysis and understand \emph{why} teams made specific reliance
decisions, an author experienced in trivia tournaments coded
decision rationales from tournament video recordings,
noting which artifacts (confidence scores, explanations,
model agreement), if any, teams cited during deliberation.
Two independent judges validated answer correctness with
95\%+ agreement, and a third expert resolved disagreements.
For evaluating the correctness of \ai{} answers, we use \abr{pedants}' judgements~\cite{li-etal-2024-pedants}.

When revising guesses with \ai{} assistance (Table~\ref{tab:collabqa-selection_method}),
humans often follow \ai{} agreement (54.8\% of decisions), followed by their own domain knowledge
(35.0\%).
Model explanations (4.4\%), confidence scores (2.2\%), and model reputation (2.0\%; ``trust
\RodeRunner{}, not \Magicarp{}, bro'')
also guide choices, though some decisions appear random (1.5\%).
Following \ai{} agreement achieves perfect accuracy (100\%), and domain knowledge performs
well (92.4\%), but confidence scores (52.3\%) perform barely better than chance, highlighting
the need for better cross-model calibration.
\vspace{-.5cm}
\begin{designprinciple}{2}{Standardize confidence across models.}
\small Systems deploying multiple \ai{} agents should invest
in cross-model calibration---especially for disagreement
cases where users need the most help.
\end{designprinciple}


\begin{table}[t]
\small
\rowcolors{2}{gray!25}{white}
\begin{tabular}{p{5cm}r}
\toprule
\textbf{Metric} & \textbf{Freq (\%)}  \\
\midrule
\multicolumn{2}{l}{\textit{Accuracy}} \\
\quad Human initial guess correct & \bonusAccHuman{}  \\
\quad\ai{} correct (random pick of \ai{}) & \bonusAccAIRandom{} \\
\quad Oracle \ai{} selection & \bonusAccAIOracle{}  \\
\quad Any human or \ai{} had correct answer & \bonusAccBestPick{}  \\
\quad Final consensus answer correct & \textbf{\bonusAccFinal{}}  \\
\midrule
\multicolumn{2}{l}{\textit{Effectiveness}} \\
\quad Humans retain correct answer & 98.0  \\
\quad Humans adopt a correct \ai{} answer & 94.4  \\
\quad Humans don't know correct answer but discern which correct \ai{}
  guess to trust & 83.3  \\
\midrule
\multicolumn{2}{l}{\textit{Failure Modes}} \\
\quad Over-reliance: humans reject their correct answer for wrong \ai{} answer & \pctOverReliance{}  \\
\quad Under-reliance: humans are wrong and fail to adopt a correct \ai{}
                answer & \pctUnderReliance{} \\
\bottomrule
\end{tabular}
\caption{How well human and \ai{} teammates work together on
  collaborative question answering.  The accuracy of the final consensus answer is
  higher than humans or computers alone and better than just picking
  the best \ai{} (which is a hard task).  However, the collaboration
  is not perfect: human teams sometimes reject their own correct
  answers or fail to adopt \ai{} answers.}
\label{tab:collabqa-bonus-metrics}
\end{table}

\subsubsection{Adoption Patterns}

\noindent\textbf{Humans often adopt \ai{} answers.}
Humans adopted an \ai{} answer 96.3\% of the time they had
no initial answer or not confident (50.7\% of cases). 
The \ai{} was correct in 73.4\%
of these cases.  The title for this paper came from \teamonlinefive{},
which was fond of saying ``\abr{ai}, take the wheel'', referencing \citet{underwood-05} as it adopted an \ai{} answer.

\begin{figure}[t]
    \centering
    \includegraphics[width=1.0\columnwidth]{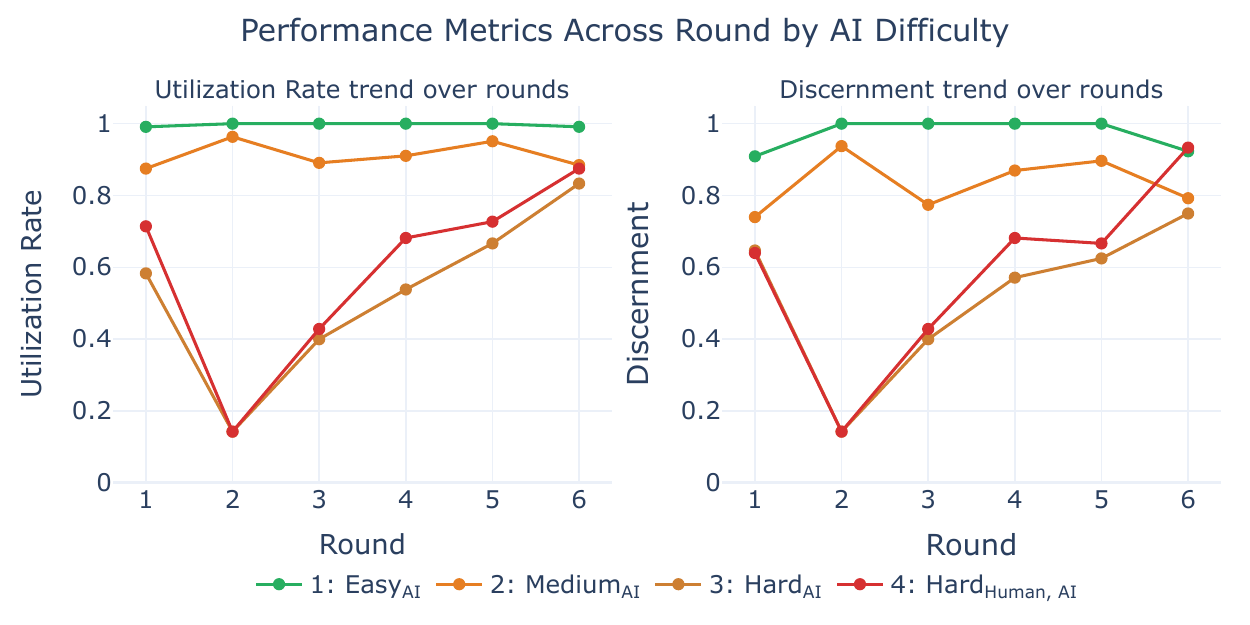}
    \caption{Utilization and discernment rates by round and question difficulty.
    As the tournament progressed, teams increasingly adopted correct \ai{} answers when available (utilization rate) and became better at selecting the right one when models disagreed (discernment),
    especially on harder questions.
    This trend reveals teams learning to use \ai{} help more effectively over time.}
    \label{fig:efficiency-trends-by-round-number}
\end{figure}

\noindent\textbf{Teams built informal \ai{} reputation through
observation alone.} Despite receiving no prior performance
information, teams' drafting choices correlated with actual
\ai{} performance (Table~\ref{tab:elo-correlation}).

\begin{table}[t]
\centering
\small
\begin{tabular}{lcc}
\toprule
\textbf{Metric} & \textbf{In-person} & \textbf{Online} \\
\midrule
Tossup Points     & +0.81*  & +0.60 \\
Bonus Points      & +0.74*  & +0.43 \\
Combined Points   & +0.76*  & +0.57 \\
Buzz Accuracy     & +0.79*  & +0.67\dag \\
1st-Buzz Accuracy & +0.67\dag & +0.71* \\
\bottomrule
\multicolumn{3}{l}{\footnotesize *$p<0.05$; \dag$p<0.10$}
\end{tabular}
\caption{Spearman $\rho$ between draft Elo ratings (from
teams' selections) and \ai{} performance metrics ($n=8$
systems per tournament). Teams drafted roughly in performance
order, especially in person, where there was more informal discussion of \ai{} properties.}
\label{tab:elo-correlation}
\end{table}

\noindent\textbf{Teams learned to use \ai{} more effectively.}
Both utilization rate (adopting available correct \ai{} answers)
and discernment (picking the right \ai{} answer when models disagreed)
increased significantly across rounds,
particularly on the hardest questions~(Figure~\ref{fig:efficiency-trends-by-round-number}; $\beta=\statTemporalBeta{}$, $p<\statTemporalP{}$).
%
This rules out simple defaulting under uncertainty: if teams
were blindly deferring to \ai{} when they had no answer,
discernment would remain near 50\%. Instead, discernment
improved from \statDiscernEarly{}\% to \statDiscernLate{}\%,
indicating that teams learned to distinguish good from bad
\ai{} responses.

\begin{designprinciple}{3}{Surface accumulating collaboration
evidence.}
\small Show where \ai{} has succeeded and failed by domain
during collaboration, rather than providing only static,
pre-deployment proficiency
summaries~\cite{yin2019understanding}.
\end{designprinciple}

\begin{figure}[t]
    \centering
    \includegraphics[width=\columnwidth]{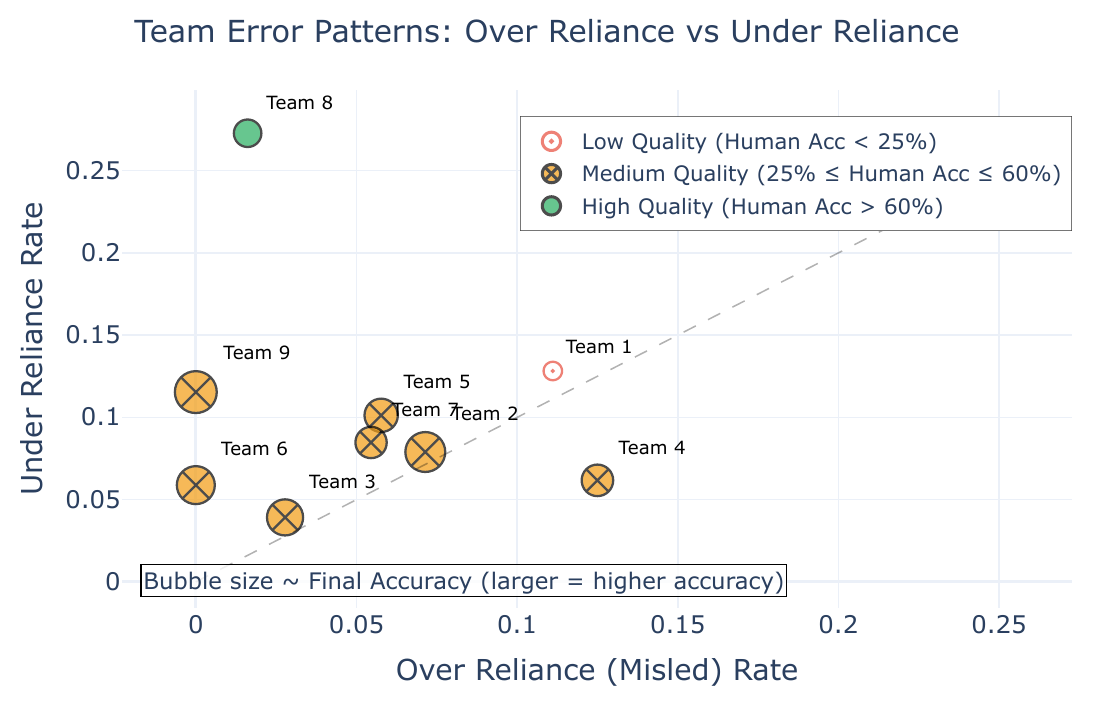}
    \caption{Team skill at question answering (marker color and shape) does not predict skill at using
    \ai{} (marker size). Middling teams had the best under/over-reliance tradeoff;
    strong teams like \teamonlinefour{} under-relied on \ai{}, while weak teams
    like \teaminpersfour{} over-relied.}
    \label{fig:bonus-team-error-scatter}
\end{figure}

\noindent\textbf{Humans overruled \ai{}, but rarely.}
Only 7.7\% of final answers differed from both \ai{} responses. This was
more likely when humans were highly confident (44.4\%), though even then
they kept their own answer only 38.1\% of the time---often revising using
\ai{}-provided content~(Figure~\ref{fig:bonus-team-error-scatter}).

\noindent\textbf{Synergy.}
With an overall accuracy of \bonusAccFinal{}\%, the overall team
(Table~\ref{tab:collabqa-bonus-metrics}) is better than the humans
alone (who get only slightly more than a third of the answers
correct; McNemar's $\chi^2=\statSynergyMcNemar{}$,
$p<\statSynergyP{}$), better than a randomly chosen \ai{} (a
little over half), and better than an oracle pick of the best \ai{}
on any given team for that specific question (three quarters of the
questions right). This shows that collaboration is mostly working as intended.

\subsubsection{Measuring Appropriate Reliance}

However, despite those promising results, we are still not at optimal
\ai{} usage.
Following \citet{schemmer2023appropriate}, we examine the error cases
of \textbf{under-reliance} and \textbf{over-reliance}.
Under-reliance is humans failing to adopt correct \ai{} advice, while
over-reliance as correct humans adopting incorrect \ai{}
advice.\footnote{Formal definitions in Appendix~\ref{appendix:collabqa-bonus-notation}.}
Both optimal rates are 0\%, although it requires both well-calibrated
\ai{} systems and discerning human users.\footnote{And it is not realistic to actually reach 0\%, as errors from incorrect / poorly written questions, out of date information, etc.\ can cause human and computer miscalibration.}
%

\jbgcomment{I think ``Good Decision'' needs to be defined: 

However, \ai{} correctness isn't binary: one \ai{} can be correct, the other wrong, or the explanation can provide a clue even if not the correct answer.  Thus, we also define \dots
}

\begin{figure}[t]
    \centering
    \includegraphics[width=\columnwidth]{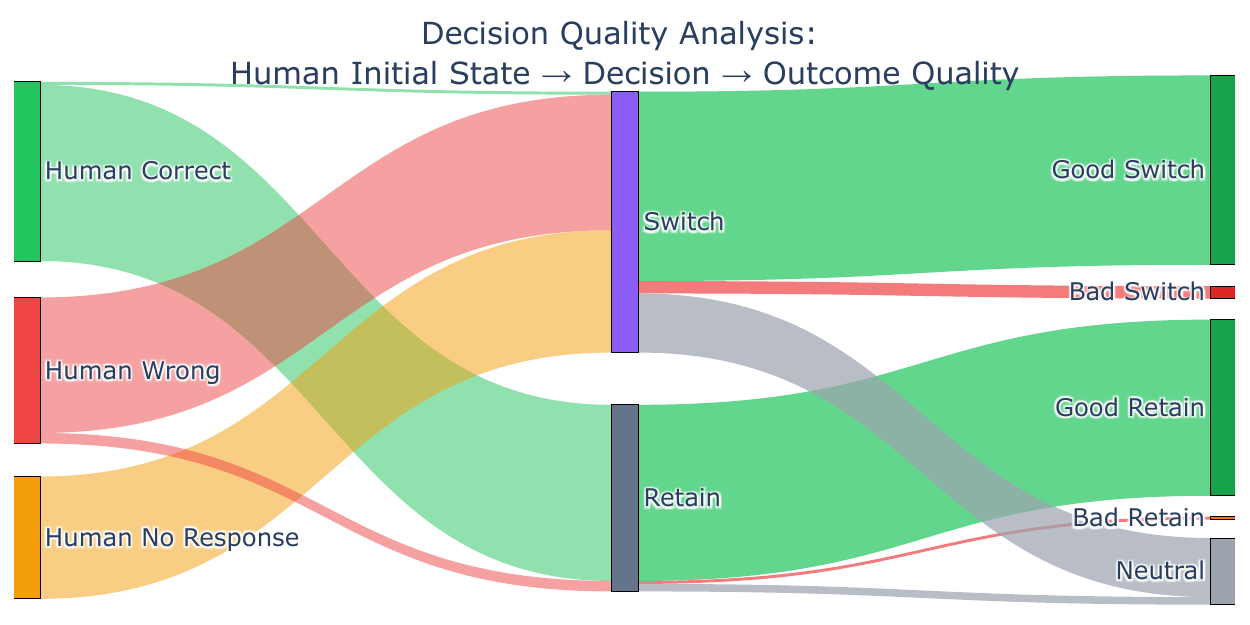}
    \vspace{-.7cm}
    \caption{Every bonus decision traced from left to right:
    whether the human's initial guess was correct (left),
    whether the team switched to an \ai{} answer (center),
    and the final outcome (right). The thickest
    flow---``Human Wrong'' through ``Switch'' to ``Good
    Switch''---represents successful collaboration: teams
    recognized their error and adopted a correct \ai{}
    suggestion. The thinner ``Human Right'' to ``Switch'' to
    ``Bad Switch'' flow represents over-reliance: teams
    abandoned a correct answer for an incorrect \ai{}
    suggestion.}
    \jbgcomment{``good decision'' needs to be explained}
    \jbgcomment{Many people are red/green colorblind.  Combine with a pattern}
    \label{fig:decision-sankey-overall}
\end{figure}

Reliance errors are asymmetric~(Figure~\ref{fig:decision-sankey-overall}): under-reliance
(\pctUnderReliance{}\% of help opportunities missed) exceeds
over-reliance (misled \pctOverReliance{}\% of times humans were
initially correct).
Teams are appropriately cautious but sometimes overly conservative.

\jbgcomment{This references qualitative findings, but doesn't connect with it.  Forward point to where those are collected.}


\begin{designprinciple}{4}{Design for mutual coverage,
targeting under-reliance.}
\small Highlight cases where \ai{} confidence is high on
domains where the user historically struggles, helping
experts recognize when to
defer~\cite{kleinberg2018human}.
\end{designprinciple}

\noindent\textbf{Consensus as a Trust Signal}
Teams interact with two \ai{}s per game, observing when models agree or
disagree.
When both \ai{}s give the same correct answer, teams switch
\pctAIConsensusSwitch{}\% of the time---above the
\pctSwitchWhenWrong{}\% average
($\chi^2=\statConsensusChi{}$, $p<\statConsensusP{}$).
When models disagree, switching drops to 45\% even when one is
correct. Consensus acts as a strong reliability signal;
disagreement signals uncertainty.

\noindent\textbf{Confirmation bias amplifies errors.}
When an incorrect human answer is confirmed by one \ai{},
under-reliance rises to
\pctUnderRelianceHaiC{}\%---teams trust the agreeing
\ai{} even when wrong
($\chi^2=\statConfirmBiasChi{}$, $p<\statConfirmBiasP{}$).
Conversely, when \ai{}s agree on an incorrect answer,
over-reliance spikes to >10\% as teams abandon correct initial guesses.
%

\begin{figure}[t]
    \centering
    \includegraphics[width=\columnwidth]{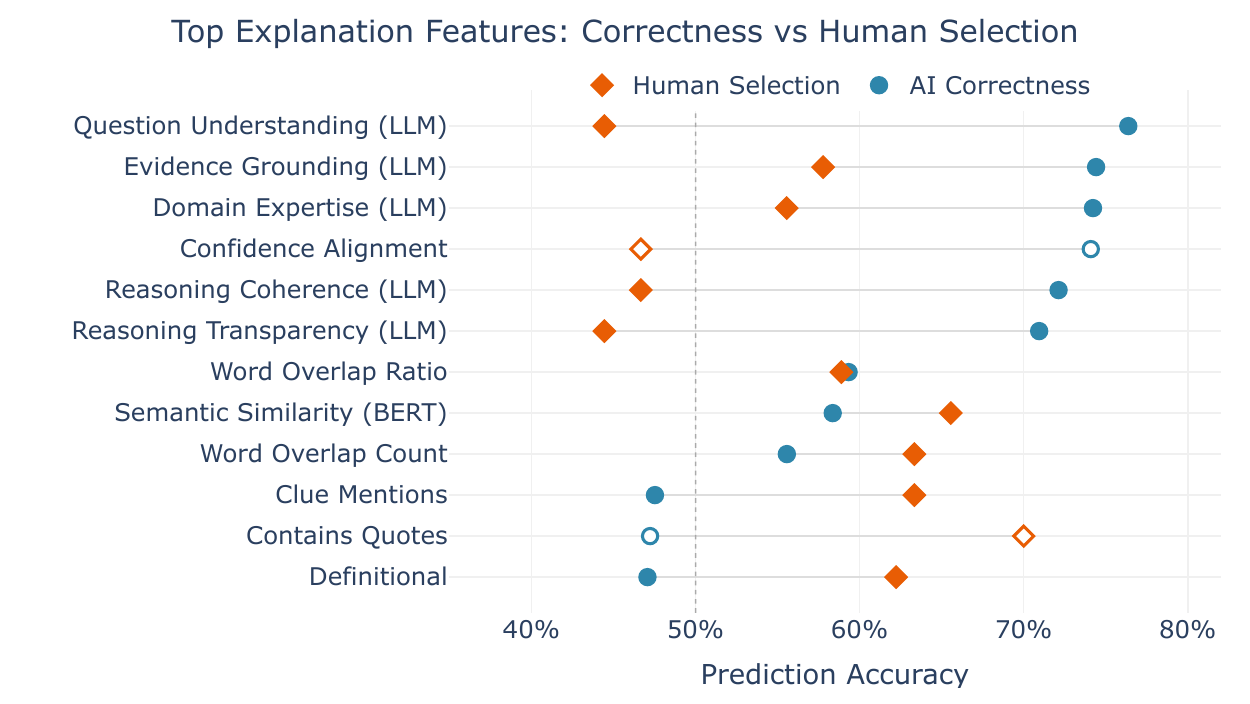}
    \caption{Explanation features differ in what predicts \ai{} correctness
    versus what humans trust. Each row shows a feature's single-predictor 
    accuracy. Filled markers indicate positive predictors (higher feature = more likely):
    hollow markers indicate negative. LLM-assessed features (Question Understanding
    76\%, Reasoning Coherence 72\%) predict correctness but humans rely on
    surface signals (quotes 70\%, \bert{} similarity 66\%). Only Evidence
    Grounding appears in both top-6 lists.}
    \label{fig:explanation-dumbbell}
\end{figure}

\noindent\textbf{What Makes Humans Trust an Explanation?}
Beyond consensus and confidence, what features of an explanation lead humans to
trust it? 
We analyze which explanation features predict adoption
decisions using single-feature logistic regressions with
significance
testing~(Section~\ref{subsec:collabqa-bonus-analysis}).
For each question, we extract 57 features from each \ai{}
explanation, spanning surface properties (length, readability),
structural patterns (quotes, clue mentions), and reasoning quality
(Appendix~\ref{appendix:collabqa-explanation-features}).
We then ask: which features predict (1) whether the \ai{} answer
is actually correct, and (2) whether humans select that
explanation?
For each feature, we fit a single-feature logistic regression
predicting either \ai{} correctness or human selection.
Features whose coefficient reaches statistical significance
($p < 0.05$) are shown with filled markers in
Figure~\ref{fig:explanation-dumbbell}; hollow markers indicate
features that did not reach significance.
Marker position shows the single-predictor accuracy: how well
that feature alone distinguishes correct from incorrect \ai{}
answers (blue) or selected from unselected explanations (orange).
\llm-assessed features like Question Understanding (76\%), Reasoning Coherence
(72\%) strongly predict correctness (Figure~\ref{fig:explanation-dumbbell}), while humans rely on surface
signals like quotes (70\%), semantic similarity (66\%), and word overlap (63\%).
\jbgcomment{Convert these to System~1 to System~N to match Table 13}
Only Evidence Grounding appears in both top lists. 
This is demonstrated by a team's comment in Round 4 of the online tournament.
On bonus 13, part 3, Magicarp gave the same guess as the players, with 95\% confidence.
\RodeRunner{} gave a different guess with 80\% confidence. 
However, players read the explanations and one stated ``I trust [\RodeRunner's] citation more''
because it was grounded in the actual bonus question text, unlike \Magicarp's.
The humans switched their guess to \RodeRunner's, which turned out to be correct.
%
%

This gap suggests a path forward: \ai{} systems should make reasoning
explicit through evidence citation, while humans should evaluate whether
explanations demonstrate genuine understanding, not just surface
familiarity.

\begin{designprinciple}{5}{Anchor explanations in evidence.}
\small Reference observable input features (specific clues,
quotes) rather than abstract reasoning. Only evidence
grounding predicts both \ai{} correctness and human
trust~\cite{vasconcelos2023explanations}.
\end{designprinciple}

%

\section{Related Work}
\label{sec:collabqa-related}
Human-\ai{} collaboration is pervasive and touches on many fields from \abr{ai} to psychology to human--computer interaction. 
Here, we focus on expertise and confidence (Appendix~\ref{appendix:collabqa-related} details related work). 

\noindent\textbf{Expertise.}
For instance, field studies analyze consequential decisions but lack
controlled interaction: observational data from judges or physicians~\cite{kleinberg2018human, gaube2021physicians} reveal reliance patterns
\jbgcomment{factor citations: I'm assuming one is judges the other is doctors}
but without behavioral information (e.g., time spent on explanations, confidence
comparisons, deliberation sequences) needed to understand decision mechanisms.
In addition, skilled practitioners are a crucial group in this context, but lab studies often do not engage genuine expertise, e.g.,  \citet{bucinca2020proxy} uses
crowdworkers to evaluate unfamiliar domains, which provides limited insight into how
experts integrate AI into practiced workflows.
\textbf{\emph{Our contribution:}} We bridge these gaps by studying
experts in their domain of practice, capturing both proactive
delegation (deciding whether to let \ai{} act before seeing its
output) and deliberative adoption (deciding whether to follow
\ai{} after evaluating its answer, confidence, and explanation).
This dual approach reveals how humans weigh evidence, calibrate
trust, and integrate \ai{} assistance.
Our setting reveals both systematic under-reliance (\pctUnderReliance{}\% 
missed opportunities) and over-reliance (when \ai{} misleads them, \pctOverReliance{}\% 
of opportunities), showing sophisticated but imperfect calibration 
that improves.

\noindent\textbf{Confidence.}
Previous work has also found that certainty/calibration \cite{sendak2020presenting}, explainability \cite{ribeiro2016should}, and  interpretability \cite{poursabzi2021manipulating} affect how much users trust an \ai{}, with both positive and negative results.
\textbf{\emph{Our contribution:}} Confidence was not a strong predictor of switching
decisions, but explanation quality modulates this effect substantially.
Specifically, explanations referencing specific evidence from the question increase appropriate switching by 
\pctExplanationImprovement{}\%.

\section{Conclusion and Discussion}

Human--\ai{} collaboration is already happening, both in proactive
delegation and deliberative adoption settings.
To improve outcomes and fully use human and \ai{} resources,
we need systems that are better calibrated, offer clear evidence,
and give people the information they need to make informed decisions.
While our study focused on text-only settings, where \ai{} agents surpassed the human teams, the synergistic collaboration exceeded the sum of its parts. This is more important in multimodal domains where \ai{} accuracy still lags.
More important will be not just facilitating collaboration with
a fixed set of team members but facilitating team
\emph{formation}: deciding which humans and which \ai{}
contributors to tap for a given problem. To support future
work, we release our dataset, tournament platform code, and
analysis scripts (Appendix~\ref{appendix:collabqa-software}).

\clearpage

\section{Limitations}

Several limitations warrant discussion:

\textbf{Domain specificity.} While our cooperative trivia tournament provides
excellent experimental control and genuine expertise, generalizing findings to
other domains requires caution.
The competitive, knowledge-intensive nature of quiz bowl may not reflect
collaborative contexts like medical diagnosis or legal review where different
decision pressures apply.
Future work should validate whether under-reliance patterns, confidence
sensitivity, and explanation effects generalize to other expert domains.


\textbf{Sample size and statistical power.} With \numHumanPlayers{} human 
players and \numAIAgents{} \ai{} agents across \numGames{} games, our study 
provides reasonable statistical power for main effects but limited ability to 
detect nuanced individual differences or rare interaction patterns.
Larger studies could reveal additional player archetypes, more precise learning
trajectories, or context-specific reliance strategies.

\textbf{Causality and interventions.} Our observational design reveals
correlational patterns between features (confidence, explanations) and reliance
decisions, but cannot establish causality.
Our annotator observes live video recordings of play and notes which artifacts
(confidence scores, explanations, model agreement) teams use to arrive at
decisions, but we cannot rule out confounding: high-confidence AI may also
generate better explanations or answer easier questions.
Future work should use randomized interventions---varying confidence levels or
explanation quality experimentally---to establish causal effects.

\textbf{Muting interface effects.} The muting mechanism required explicit action
before each toss-up, potentially introducing friction that inflates muting
rates.
Alternative designs (default-on with quick toggle, voice commands) might yield
different strategic patterns.
However, the systematic learning and context-dependence we observe suggests
humans engaged strategically rather than randomly.

\textbf{Question design and ecological validity.} Our adversarial questions
intentionally exploit known AI weaknesses, creating systematic skill gaps.
While this validates complementarity opportunities, real-world collaborative
contexts may feature less predictable AI failure modes.
However, the design principles we derive---prioritizing calibrated confidence,
grounded explanations, and user control---should generalize beyond adversarial
settings.

\textbf{Temporal scope.} We observe learning effects across four tournament
rounds (approximately 2 hours per session), but cannot assess long-term trust
calibration.
Extended collaboration might reveal different patterns: humans could become
overconfident in AI after positive experiences, or develop more sophisticated
mental models enabling better calibration.
Longitudinal studies tracking trust evolution over weeks or months would
complement our findings.
\section{Ethics Statement}
The experiments performed in this study involved
human participants. All the experiments involving human evaluation in this paper were exempt under institutional IRB review. 

All human data collection procedures were reviewed and approved by an institutional review board (IRB) to ensure the protection of participants’ privacy and rights. Human buzzpoints in the dataset are fully anonymized. Although the post-competition survey collected participant names for compensation purposes, only aggregate statistics and anonymized quotes are reported in the study.

Trivia players collectively received \$600 USD in online gift cards as prizes for the competitions, with awards of \$150, \$100, and \$50 for the top three teams in both offline and online tournaments. Model submitters received a total of \$400 USD in online gift cards, distributed as \$200, \$150, \$100, and \$50 prizes. Question writers were compensated \$5 per question, and editors \$1 per edited question, corresponding to an estimated rate exceeding \$10 USD per hour—above the U.S. federal minimum wage of \$7.50 USD.
All the involved participants gave their consent to disclose their interactions with the interface.
The documents used in the study are distributed under an open license.

\textbf{IRB approval and informed consent.} This study was approved by our institutional review board (IRB) to ensure participant privacy and rights. All participants gave informed consent after being clearly told the study involved human-AI collaboration and behavioral data collection. Participation was voluntary and could be withdrawn at any time without penalty. The IRB monitored procedures for collecting responses and questions to protect privacy throughout the study.

\textbf{Participant compensation and treatment.} Human players participated
voluntarily, motivated by competitive interest in quiz bowl rather than
monetary compensation (consistent with standard quiz bowl tournament practice).
Question writers received fair compensation (\$\writerCompensation{}/hour, 
above typical rates for quiz bowl writing).
All participants were treated with respect throughout the study.

\textbf{Data privacy and transparency.} We collected only gameplay data
(answers, timing, muting decisions, switching decisions) and basic demographic
information (experience level).
No personally identifiable information beyond pseudonymous player IDs appears
in our dataset or analysis.
Participants were informed about AI system characteristics (model types, skill
levels) but not specific implementation details that might affect strategic
behavior.

\textbf{AI attribution and disclosure.} Participants knew they were
collaborating with AI agents and understood when AI suggestions appeared.
We clearly disclosed confidence scores and explanations as AI-generated rather
than human expert opinions.
No deception was used; all AI interactions were transparent.

\textbf{Community engagement and data release.} We engaged
with the quizbowl community throughout this research, explaining
our goals and design choices to tournament organizers and
participants. We release the full dataset
(questions, responses, behavioral traces), the tournament web
application source code, and all analysis scripts to support
reproducibility. See
Appendix~\ref{appendix:collabqa-software} for platform details.

\textbf{Broader impacts.} This research advances understanding of human-AI
collaboration with potential benefits for interface design in high-stakes
domains (medical diagnosis, legal review).
However, our findings about systematic under-reliance could be misused to
encourage blind trust in AI systems.
We emphasize that appropriate reliance requires well-calibrated AI confidence
and grounded explanations---trust should be conditional on AI system quality,
not automatic.

\textbf{Use of AI assistants.} The authors used AI tools (OpenAI's ChatGPT
and Anthropic's Claude) for coding assistance during data analysis and
visualization, and as a writing assistant limited to paraphrasing for
conciseness.
All substantive content, analysis, and conclusions are the authors' own work.





\section*{Acknowledgements}

We thank the UMD CLIP group for valuable feedback
on study design and adversarial evaluation. We are grateful to all
33 participants who developed and submitted \ai{} systems or played
in the live trivia tournament. We also thank our question writers
and editors for their contributions. In particular, we thank those who
agreed to be named: Neel Mokaria and Amanvir Parhar for their model
submissions, Ankit Aggarwal and Kartik Ravisankar for their question
authoring, and Jaimie Carlson for both submitting a model and writing
questions. We appreciate all who made the experiment possible.

This work was supported in part by the National Science Foundation (IIS-2403436)
and the NSF Institute for Trustworthy AI in Law \& Society (TRAILS, 2229885).
Any opinions, findings, conclusions, or recommendations expressed are those of the
authors and do not necessarily reflect the views of the sponsors.

\bibliography{bib/journal-full,bib/custom,bib/humanai,bib/jbg}

\appendix

\section{Additional Related Work}
\label{appendix:collabqa-related}

This section outlines additional related work not featured in the main text of the paper.

\jbgcomment{In general, this should be doing more contextualization / comparison with our own work.}

\jbgcomment{Add signposting of what's going to be covered}

\subsection{Human-AI Collaboration and Appropriate Reliance}

Understanding appropriate reliance in automation has long been central to human
factors research \cite{parasuraman1997humans, lee2004trust}.
\jbgcomment{Don't cite twice in the same paragraph, just use later cites.}
Parasuraman and Riley's foundational taxonomy distinguishes misuse
(over-reliance on unreliable automation) from disuse (under-reliance on
reliable automation), with trust calibration---matching reliance to actual
capability---identified as crucial for effective human-automation teams
\cite{lee2004trust}.
\jbgcomment{``identified as crucial'' is awkward phrasing.  Why is it actually important}
\jbgcomment{trust dimensions needs to be defined / explained}
Early work focused on measuring trust dimensions \cite{madsen2000measuring,
hoff2015trust}, establishing that trust must align with system capability for
optimal collaboration.

Recent work demonstrates that mental model accuracy---not just AI accuracy---predicts
team performance.
\citet{bansal2019beyond} show that when humans develop accurate understanding of AI
strengths and weaknesses, team performance improves even when AI accuracy
remains constant.
However, explanations can paradoxically reduce complementarity by inducing
over-reliance: teams exposed to AI explanations sometimes defer blindly rather
than critically evaluating suggestions \cite{bansal2021does}.
Other work formalizes the learning-to-complement problem, developing
algorithms that optimize for team performance rather than individual AI
accuracy \cite{wilder2021learning}.
Steyvers et al. provide a Bayesian framework for human-AI complementarity,
showing how optimal aggregation depends on confidence calibration and skill
correlation \cite{steyvers2022bayesian}.
Zhang et al. find that confidence scores and explanations have complex
interactions: high confidence can increase trust, but explanations sometimes
fail to improve calibration \cite{zhang2020effect}.

Field studies reveal persistent reliance problems across domains.
Judges systematically ignore helpful AI bail recommendations, suggesting
algorithmic aversion or mistrust \cite{kleinberg2018human}.
In contrast, physicians over-rely on AI in medical imaging even when
initially correct, changing correct diagnoses to match AI suggestions
\cite{gaube2021physicians}.
A systematic review identifies under-reliance as more common than
over-reliance across human-AI systems, though context matters substantially
\cite{hemmer2021human}.
\citet{green2019disparate} show that users adapt AI use strategically across contexts,
suggesting reliance patterns are neither fixed nor random but responsive to
perceived task demands.

Despite this progress, three methodological gaps limit our understanding of
reliance mechanisms.
\textit{First}, most controlled studies measure post-hoc acceptance rates after
humans see AI suggestions \cite{bansal2021does, zhang2020effect}, capturing
whether people accept advice but not \textit{how} they decide---the real-time
evaluation process, information weighting, or decision criteria remain opaque.
\textit{Second}, field studies analyze consequential decisions but lack
controlled interaction: observational data from judges or physicians
\cite{kleinberg2018human, gaube2021physicians} reveal reliance patterns
without the behavioral traces (e.g., time spent on explanations, confidence
comparisons, deliberation sequences) needed to understand decision mechanisms.
\textit{Third}, lab studies achieving experimental control often use synthetic
tasks that fail to engage genuine expertise \cite{bucinca2020proxy}---
crowdworkers evaluating unfamiliar domains provide limited insight into how
experts integrate AI into practiced workflows.

\textbf{Our contribution:} We bridge these gaps by studying experts in their
domain of practice, capturing both proactive strategic decisions (muting before
seeing AI output) and deliberative evidence-based decisions (switching after
evaluating AI's answer, confidence, and explanation).
This dual-signal approach reveals the reliance process itself: how humans weigh
evidence, calibrate trust, and integrate AI assistance.
Our setting reveals both systematic under-reliance (\pctUnderReliance{}\% 
missed opportunities) and over-reliance (misled on \pctOverReliance{}\% 
of opportunities), showing sophisticated but imperfect calibration 
that improves.

\subsection{Trust Calibration: Confidence and Explanations}

Appropriate reliance requires well-calibrated AI confidence.
Modern neural networks are poorly calibrated: high-confidence predictions
often prove incorrect, while low-confidence predictions can be accurate
\cite{guo2017calibration}.
Structured presentation of uncertainty—communicating both point estimates and
confidence intervals—helps users make better decisions than point estimates
alone \cite{bhatt2021uncertainty}.
In medical domains, confidence-aware AI assistance improves diagnostic
accuracy compared to AI without confidence scores \cite{tschandl2020human}.
Model facts labels that present calibration information to clinical users
improve trust calibration \cite{sendak2020presenting}, and uncertainty
quantification can reduce overreliance when implemented carefully
\cite{marusich2024using}.

Explanations have complex, sometimes counterintuitive effects on reliance.
While explainability methods like LIME \cite{ribeiro2016should} aim to
increase transparency, explanations can manipulate trust through plausible
but incorrect rationales \cite{lakkaraju2020fool}.
\jbgcomment{Use citet in cases like this}
Poursabzi-Sangdeh et al. find that greater interpretability can increase
overconfidence rather than improving calibration \cite{poursabzi2021manipulating}.
More promisingly, cognitive forcing functions—interventions requiring
engagement with explanations before making decisions—reduce over-reliance by
prompting critical evaluation \cite{bucinca2021trust}.
Vasconcelos et al. demonstrate that explanations help when they reveal AI
limitations: showing where and why AI might fail improves reliance decisions
more than generic explanations of reasoning \cite{vasconcelos2023explanations}.
The type of explanation matters: example-based explanations affect trust
differently than feature importance \cite{cai2019effects}, and social
transparency (explaining system context and design choices) complements
technical explanations \cite{ehsan2021expanding}.

\textbf{Our contribution:} We jointly analyze confidence and explanations in
naturalistic decisions, finding that grounded explanations---those referencing
specific evidence from the question---increase appropriate switching by 
\pctExplanationImprovement{}\%, providing actionable design guidance beyond 
generic explanation requirements.
We show that confidence was not a strong predictor of switching
decisions, but explanation quality modulates this effect substantially. 
Moreover, two AI companions are not always better than one, and the consensus among AIs is not always helpful.

\subsection{Adversarial Evaluation and Question Answering}

Standard QA datasets like SQuAD \cite{rajpurkar2016squad, rajpurkar2018know}
measure AI progress but often saturate as models exploit superficial
patterns.
Adversarial evaluation exposes systematic weaknesses: adding distracting
sentences with keyword overlap breaks reading comprehension models
\cite{jia2017adversarial}, and human-in-the-loop adversarial generation
creates natural-seeming questions that fool state-of-the-art systems
\cite{wallace2019trick}.
Contrast sets—minimal edits changing correct answers—reveal that models rely
on spurious correlations rather than robust reasoning
\cite{gardner2020evaluating}.
Shortcut learning explains these failures: neural networks exploit dataset
artifacts rather than learning intended capabilities \cite{geirhos2020shortcut}.

Quiz bowl provides a compelling testbed for incremental question answering.
\citet{boyd2012besting} introduced the QANTA dataset with pyramidal questions
where clues progress from obscure to obvious.
\citet{rodriguez2019quizbowl} formalize buzz timing as a confidence signal: systems must
decide when to answer based on expected utility, balancing accuracy against
competitive risk.
Recent work systematically generates context-aware adversarial examples that
exploit model weaknesses while maintaining naturalness \cite{sung-etal-2025-benchmark}.

\textbf{Our contribution:} While adversarial NLP typically focuses on AI
failure, we use adversarial design to create complementarity opportunities—
questions where human-AI skill gaps enable collaboration gains.
This reframes adversarial evaluation from exposing weaknesses to engineering
productive partnerships: AI-favoring questions create under-reliance
reduction opportunities, while human-favoring questions test appropriate
skepticism.

\subsection{Strategic Behavior and Expertise}

Cognitive science distinguishes expert intuition, which relies on holistic
pattern recognition developed through extensive experience \cite{klein1998sources,
chi2014cambridge}, from novice deliberation following explicit rules
\cite{dreyfus1986mind}.
Kahneman's dual process framework contrasts fast, automatic System 1 thinking
with slow, deliberative System 2 reasoning \cite{kahneman2011thinking}.
These distinctions raise questions about expert-AI integration: do experts
rely on intuition when evaluating AI, or does AI assistance shift them toward
deliberative reasoning?

Users adapt AI use strategically across contexts rather than maintaining
fixed reliance patterns \cite{green2019disparate}.
Users slowly learn to calibrate trust as they observe AI performance across
repeated interactions \cite{yin2019understanding}.
However, subjective measures like self-reported trust can mislead: objective
behavioral outcomes matter more than stated preferences
\cite{bucinca2020proxy}.
Perceived control moderates AI acceptance—users value the ability to override
AI even when they rarely exercise it \cite{vaccaro2018illusion}.

\label{appendix:collabqa-design-principles}

\begin{table}[t]
\centering
\small
\begin{tabularx}{\columnwidth}{>{\raggedright\arraybackslash}p{2.8cm}X}
\toprule
\textbf{Principle} & \textbf{Guideline} \\
\midrule
\rowcolor{gray!10}
\textbf{Calibrated confidence} &
Invest in standardized confidence scales; within-model calibration
works, but cross-model comparison fails. \\
\addlinespace[2pt]
\textbf{Grounded explanations} &
Anchor in observable input (``The clue mentions X...'') rather than
abstract reasoning; use calibrated scores over hedges. \\
\addlinespace[2pt]
\rowcolor{gray!10}
\textbf{Strategic control} &
Provide context-dependent toggles (by topic, difficulty) rather than
binary on/off; user agency over \emph{when} \ai{} participates matters. \\
\addlinespace[2pt]
\textbf{Collaboration feedback} &
Surface context-specific analytics (``\ai{} helped on 8/10 science
questions'') to accelerate trust calibration. \\
\addlinespace[2pt]
\rowcolor{gray!10}
\textbf{Expert adoption} &
Highlight \ai{} strengths on question types where users struggle;
under-reliance (\pctUnderReliance{}\%) exceeds over-reliance (\pctOverReliance{}\%). \\
\bottomrule
\end{tabularx}
\caption{Design principles for human--\ai{} collaboration systems derived
from our behavioral findings. Each guideline addresses a specific failure
mode observed in our tournaments.}
\label{tab:collabqa-design-principles}
\end{table}

Individual differences moderate reliance quality: cognitive ability predicts
susceptibility to biases \cite{stanovich2000individual}, and metacognitive
accuracy—knowing what you know—varies substantially across individuals
\cite{fleming2014measure}.
Expertise moderates how users integrate AI assistance: domain experts process
explanations differently than novices, sometimes exhibiting appropriate
skepticism that novices lack \cite{lai2019human}.

\textbf{Our contribution:} We study expert humans in a competitive domain
with genuine expertise requirements, capturing strategic reliance (costly
muting) and deliberative reliance (explanation-mediated choice).
This reveals systematic patterns: experts under-rely (missing 
\pctUnderReliance{}\% of opportunities) but over-rely (when \ai{} misleads them, \pctOverReliance{}\% 
of opportunities), showing sophisticated but imperfect calibration.
Trust improves across rounds (+\pctCalibrationImprovement{}\% from Round 1 to 
Round 7), demonstrating learnable calibration, and high-skill players show 
better calibration than low-skill players (gap=8\% vs 22\%), consistent with 
expertise moderating AI integration.
\section{\Bonus{} Phase Analysis: Extended Details}
\label{appendix:collabqa-bonus-details}

This appendix provides additional details for the deliberative adoption
analysis in Section~\ref{subsec:collabqa-bonus-analysis}.

\subsection{Formal Notation}
\label{appendix:collabqa-bonus-notation}

We formalize the bonus phase interaction as follows. After the human team
confers and provides an initial response, the final human answer $h_c \in
\{0, 1\}$ indicates correctness.\footnote{Multiple team members deliberate,
so an individual member may have the correct answer without it being given
as the final team response---this is still recorded as $h_c = 0$.}

The team then sees responses from two \ai{} agents. We define:
\begin{itemize}
    \item $a_{gc} \in \{0, 0.5, 1\}$: \ai{} guess correctness (0 if both
    wrong, 0.5 if one correct, 1 if both correct)
    \item $a_{rc} \in \{0, 0.5, 1\}$: \ai{} recall correctness, measuring
    whether the correct answer appears in either model's full output
    including explanations
    \item $s \in \{0, 1\}$: whether the team switched from their initial
    response
    \item $f_c \in \{0, 1\}$: final team correctness after deliberation
\end{itemize}

Following \citet{schemmer2023appropriate}, we define reliance as depending
on whether the \ai{} was correct and whether humans adopt it:
\begin{align}
\text{Under-reliance} &= P(s = 0 \mid h_c = 0, a_{rc} > 0) \\
\text{Over-reliance} &= P(s = 1 \mid h_c = 1, a_{gc} < 1)
\end{align}
Under-reliance captures missed help opportunities; over-reliance captures
being misled by incorrect \ai{} suggestions. Both optimal rates are 0\%.




\subsection{Selection Method Breakdown}
\label{appendix:collabqa-selection-methods}

When humans revise their response using \ai{} assistance, how do they choose
which guess to trust? Table~\ref{tab:collabqa-selection_method} breaks down
selection methods and their effectiveness. We restrict this table to revised
cases with a meaningful choice among candidate answers: either the two \ai{}
systems disagree, or the \ai{} consensus conflicts with the initial human
answer. We also require that the initial human answer or at least one \ai{}
candidate is correct, so final accuracy reflects whether the team chose a
correct available answer.
The first row reports cases where the two \ai{} systems agreed and the team
adopted that shared answer, as a share of all eligible cases. For the remaining
cases, the denominator resets to the non-agreement cases. These rows report the
relative frequency of each selection method within that remainder and the mean
final correctness for cases using that method.

\begin{table*}[!t]
    \centering
    \small
    \begin{tabular}{lccc}
    \toprule
    \textbf{Case} & \textbf{Share of} & \textbf{Share of} & \textbf{Final} \\
     & \textbf{all cases} & \textbf{disagreement cases} & \textbf{accuracy} \\
    \midrule
    AI Agreement & 55.0\% & -- & 100.0\% \\
    \addlinespace
    AI Disagreement &  &  &  \\
    \quad Domain Knowledge & 33.9\% & 75.2\% & 90.9\% \\
    \quad Model Explanation & 4.4\% & 9.8\% & 83.3\% \\
    \quad Random Selection & 3.2\% & 7.2\% & 50.0\% \\
    \quad Model Reputation & 1.9\% & 4.2\% & 69.2\% \\
    \quad Model Confidence & 1.6\% & 3.6\% & 54.5\% \\
    \bottomrule
    \end{tabular}
    \caption{Selection methods used by humans when revising their responses with
    \ai{} assistance. AI agreement is reported as a share of all eligible cases
    ($N=682$). The remaining rows report the distribution
    of selection methods among non-agreement cases ($N=307$)
    and their final accuracy.}
    \label{tab:collabqa-selection_method}
\end{table*}

\paragraph{Key Findings.}
AI agreement accounts for 55.0\% of eligible revised cases and is perfectly
accurate in this filtered set. Among the remaining cases, domain knowledge
dominates selection decisions, accounting for 75.2\% of choices with 90.9\%
final accuracy. This suggests teams are most effective when they can use their
own expertise to evaluate \ai{} suggestions rather than relying on surface
signals.
Model explanations account for 9.8\% of the remaining choices and reach
83.3\% final accuracy. Model reputation and confidence are less common
signals, at 4.2\% and 3.6\% respectively, and confidence alone has lower
accuracy (54.5\%).
Notably, 7.2\% of selections appeared random, with 50.0\% final accuracy,
suggesting that teams sometimes lacked a clear basis for choosing among
conflicting answers.

\subsection{Decision Flow by AI Agreement}
\label{appendix:collabqa-decision-sankey-full}

\begin{figure*}[p]
    \centering
    \includegraphics[width=\textwidth]{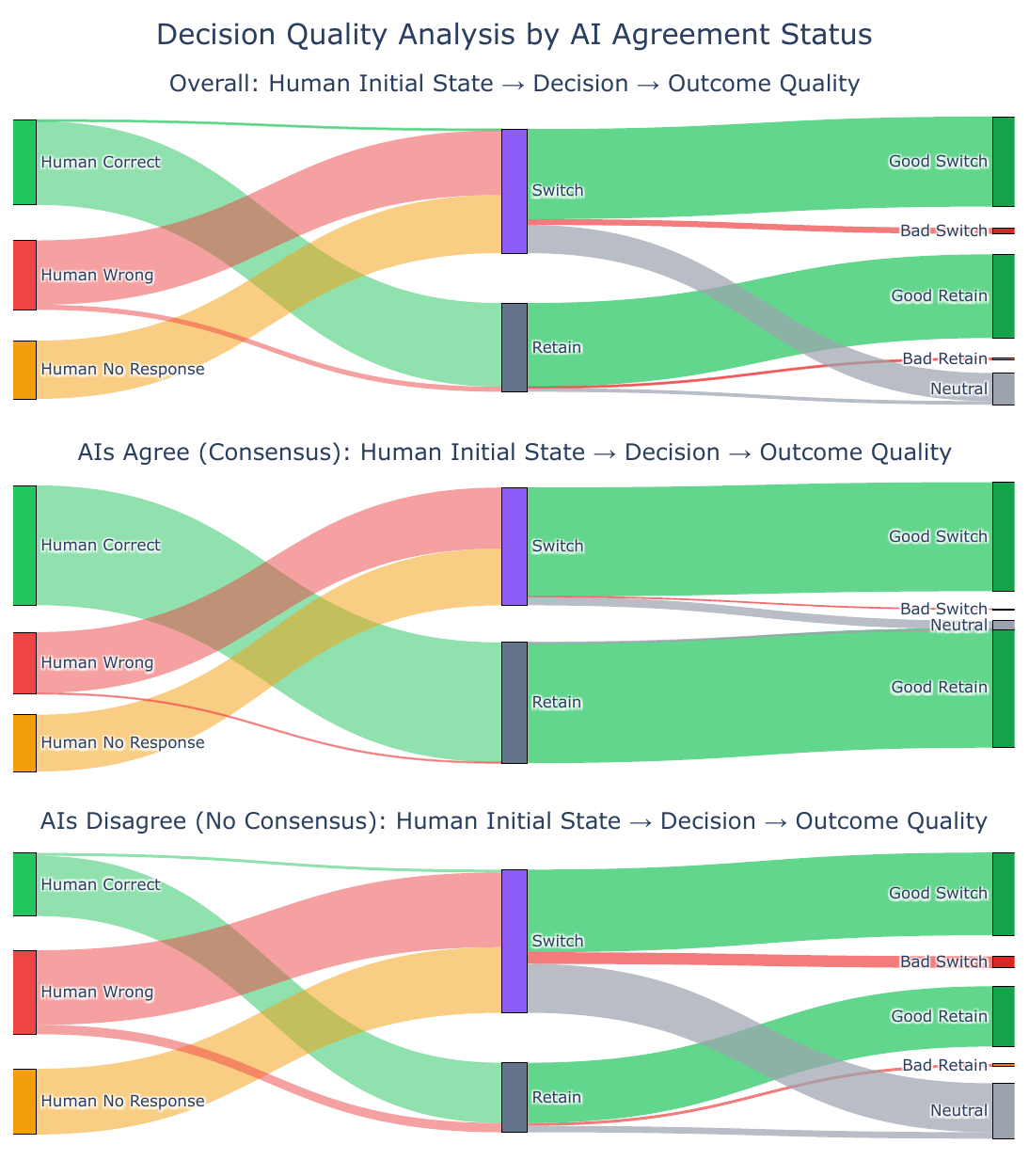}
    \caption{Decision quality analysis by \ai{} agreement status. Top: overall
    flow from human initial state through decision to outcome. Middle: when
    \ai{}s agree (consensus), teams switch more readily and achieve higher
    accuracy. Bottom: when \ai{}s disagree, teams face harder choices and
    under-reliance increases. Flow widths represent proportion of responses;
    colors indicate outcome quality. Compare to the aggregate view in
    Figure~\ref{fig:decision-sankey-overall}.}
    \label{fig:decision-sankey-full}
\end{figure*}

Figure~\ref{fig:decision-sankey-overall} in Section~\ref{subsec:collabqa-bonus-analysis}
shows the aggregate decision flow. Here we present the full breakdown
conditioned on whether the two \ai{} agents agreed
(Figure~\ref{fig:decision-sankey-full}).

\paragraph{Key Observations.}
When \ai{}s reach consensus, teams benefit from a clear signal: switching to
the agreed answer yields high accuracy, and under-reliance drops substantially.
When \ai{}s disagree, teams must evaluate competing suggestions, leading to
higher cognitive load and more conservative behavior (increased inaccuracies).
This pattern highlights the value of \ai{}-\ai{} agreement as a trust signal,
though it can also amplify errors when both \ai{}s are wrong
(Section~\ref{subsec:collabqa-bonus-analysis}).

\subsection{Reliance by AI Agreement Condition}
\label{appendix:collabqa-consensus-reliance}

\begin{table*}[t]
\centering
\small
\setlength{\tabcolsep}{5pt}
\begin{tabular}{@{}l*{5}{c}@{}}
\toprule
 & Overall & AI consensus & AI disagree & H--AI confirm & H--AI conflict \\
\midrule
\multicolumn{6}{@{}l}{\textit{Under-reliance}} \\
\textbf{Overall} & 3.9\%\,($\scriptstyle n\!=\!998$) & \cellcolor[HTML]{FFE0B2}{0.7\%\,($\scriptstyle n\!=\!417$)$^{*}$} & 6.2\%\,($\scriptstyle n\!=\!581$) & \cellcolor[HTML]{FFCC80}{64.5\%\,($\scriptstyle n\!=\!31$)$^{**}$} & 2.0\%\,($\scriptstyle n\!=\!967$) \\
\textbf{Confident} & \cellcolor[HTML]{FFCC80}{16.9\%\,($\scriptstyle n\!=\!178$)$^{**}$} & 3.1\%\,($\scriptstyle n\!=\!64$) & \cellcolor[HTML]{FFCC80}{24.6\%\,($\scriptstyle n\!=\!114$)$^{**}$} & \cellcolor[HTML]{FFCC80}{81.0\%\,($\scriptstyle n\!=\!21$)$^{**}$} & 8.3\%\,($\scriptstyle n\!=\!157$) \\
\textbf{Unsure} & \cellcolor[HTML]{FFCC80}{1.1\%\,($\scriptstyle n\!=\!820$)$^{**}$} & \cellcolor[HTML]{FFE0B2}{0.3\%\,($\scriptstyle n\!=\!353$)$^{*}$} & 1.7\%\,($\scriptstyle n\!=\!467$) & \cellcolor[HTML]{FFF3E0}{30.0\%\,($\scriptstyle n\!=\!10$)$^{\dagger}$} & \cellcolor[HTML]{FFCC80}{0.7\%\,($\scriptstyle n\!=\!810$)$^{**}$} \\
\midrule
\multicolumn{6}{@{}l}{\textit{Over-reliance}} \\
\textbf{Overall} & 1.7\%\,($\scriptstyle n\!=\!666$) & 0.0\%\,($\scriptstyle n\!=\!421$) & 4.5\%\,($\scriptstyle n\!=\!245$) & 0.5\%\,($\scriptstyle n\!=\!604$) & \cellcolor[HTML]{FFCC80}{12.9\%\,($\scriptstyle n\!=\!62$)$^{**}$} \\
\textbf{Confident} & 1.3\%\,($\scriptstyle n\!=\!632$) & 0.0\%\,($\scriptstyle n\!=\!398$) & 3.4\%\,($\scriptstyle n\!=\!234$) & 0.5\%\,($\scriptstyle n\!=\!576$) & \cellcolor[HTML]{FFF3E0}{8.9\%\,($\scriptstyle n\!=\!56$)$^{\dagger}$} \\
\textbf{Unsure} & 8.8\%\,($\scriptstyle n\!=\!34$) & 0.0\%\,($\scriptstyle n\!=\!23$) & \cellcolor[HTML]{FFE0B2}{27.3\%\,($\scriptstyle n\!=\!11$)$^{*}$} & 0.0\%\,($\scriptstyle n\!=\!28$) & \cellcolor[HTML]{FFCC80}{50.0\%\,($\scriptstyle n\!=\!6$)$^{**}$} \\
\bottomrule
\end{tabular}
\caption{Under- and over-reliance rates by human confidence
and \ai{} agreement condition. Each cell shows the error rate
and $n$ (denominator). Significance markers compare each cell
to the panel baseline (Overall row, Overall column):
$^{**}$\,Bonferroni $p<0.01$, $^{*}$\,$p<0.05$, $^{\dagger}$\,$p<0.1$.
Shading intensity matches significance.}
\label{tab:nested-reliance-consensus}
\end{table*}

Table~\ref{tab:nested-reliance-consensus} breaks down over-
and under-reliance rates across four conditions defined by
whether one of the \ai{} teammates agreed with the initial
human response. When an incorrect human answer is confirmed by
one \ai{} while the other provides the correct answer,
under-reliance spikes to \pctUnderRelianceHaiC{}\%
(\S\ref{subsec:collabqa-bonus-analysis}).




\section{Dataset Overview}
\label{appendix:collabqa-dataset-overview}

This appendix provides an overview of the dataset collected in the tournament.

\begin{table}[t]
\centering
\small
\begin{tabularx}{\columnwidth}{l>{\raggedleft\arraybackslash}p{1.5cm}X}
\toprule
\textbf{Component} & \textbf{Count} & \textbf{Description} \\
\midrule
Tournaments & \numTournaments{} & Offline (June 14) + Online (June 21) \\
Games & \numGames{} & \numTossupsPerGame{} tossups + 
\numBonusesPerGame{} bonuses each \\
Human players & \numHumanPlayers{} & Experienced quizbowl competitors \\
AI agents & \numAIAgents{} & Varying skill levels and architectures \\
Unique questions & \numTossups{} each & \numTossups{} tossups, 
\numBonusParts{} bonuses \\
Tossup responses & \numProactiveDecisions{} & With muting state, buzz timing \\
Bonus responses & \numDeliberativeDecisions{} & With AI suggestions per part \\
Muting decisions & $\sim$\numMutingDecisions{} & Strategic AI disabling \\
Switching decisions & $\sim$\numSwitchingDecisions{} & Changed answer after 
AI input \\
\bottomrule
\end{tabularx}
\caption{Dataset overview showing the scale and richness of behavioral 
data collected during human-AI collaborative quizbowl tournaments. The 
dataset captures multiple types of reliance decisions with real stakes.}
\jbgcomment{We probably don't need all of this in the main body or in a table.  Some should go to appendix.}
\label{tab:collabqa-dataset-overview}
\end{table}

\section{Our Adversarial Questions}
\label{appendix:collabqa-adversarial-questions}

We categorize each bonus question by difficulty for humans and AI separately,
based on average accuracy across all responses. Questions with accuracy below
40\% are labeled \emph{Hard}, 40--70\% as \emph{Medium}, and above 70\% as
\emph{Easy}. We then compute \emph{relative difficulty} by comparing human and
AI accuracy: questions where humans outperform AI by more than 20 percentage
points are \emph{Harder for AI}, questions where AI outperforms humans by the
same margin are \emph{Harder for Human}, questions where both struggle (below
40\% accuracy) are \emph{Hard for Both}, and the remainder are \emph{Balanced}.

\begin{figure}[!h]
\centering
\includegraphics[width=\columnwidth]{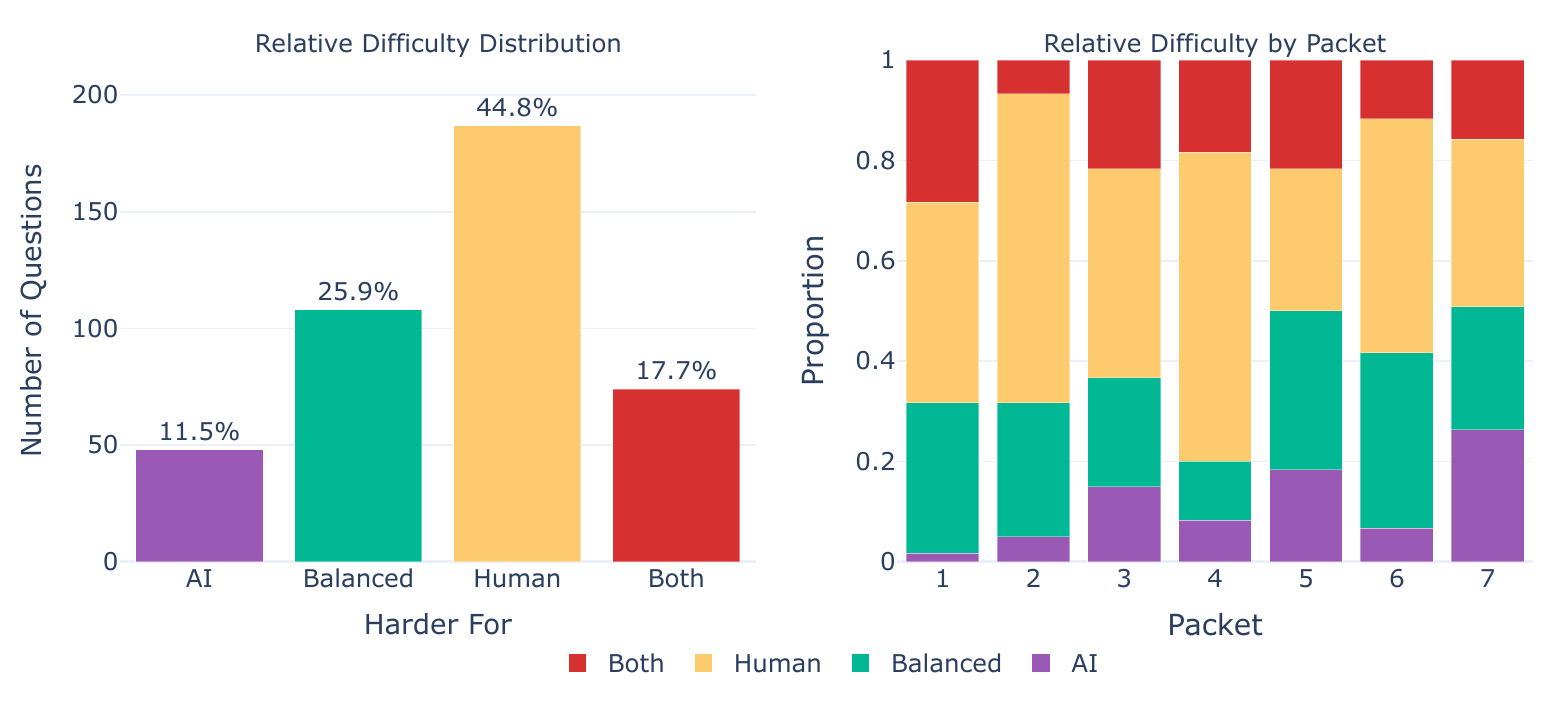}
\caption{Relative difficulty distribution across bonus questions. Left: overall
counts by category. Right: proportion breakdown by packet showing consistent
patterns across question sets.}
\label{fig:relative-difficulty}
\end{figure}

Figure~\ref{fig:relative-difficulty} shows the distribution of relative
difficulty. The largest category (45\%) comprises questions harder for humans,
where AI outperforms. About 26\% fall into the \emph{Balanced} category where
humans and AI perform comparably. Questions hard for both (18\%) represent
challenging items that neither humans nor AI answer reliably. Finally, 11\% are
harder for AI, where humans have a clear advantage. The right panel shows this
breakdown by packet, revealing variation across question sets.

\begin{figure}[!h]
\centering
\includegraphics[width=\columnwidth]{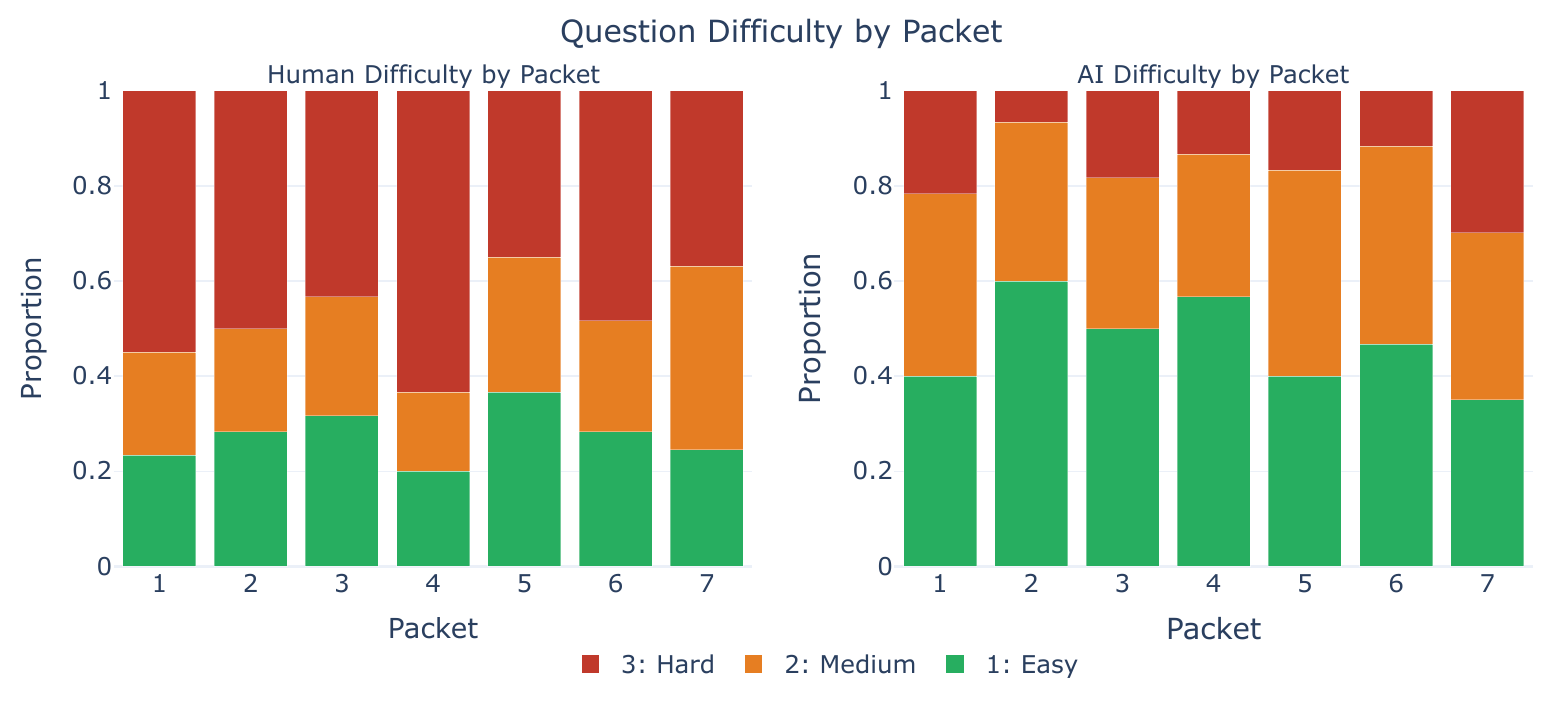}
\caption{Question difficulty by packet for humans (left) and AI (right). Each
bar shows the proportion of Easy, Medium, and Hard questions within a packet.
Human difficulty varies more across packets than AI difficulty.}
\label{fig:difficulty-by-packet}
\end{figure}

Figure~\ref{fig:difficulty-by-packet} compares absolute difficulty for humans
and AI across packets. Human difficulty shows more variation, with some packets
containing a higher proportion of hard questions. AI difficulty is more uniform
across packets, suggesting that question design affects humans and AI
differently. This variation creates diverse collaboration scenarios within each
tournament.

\section{Draft Mechanics}
\label{appendix:collabqa-draft-mechanics}

This appendix details the \draftSelection{} procedure used in our tournaments.
The draft mechanism varied slightly between the in-person and online tournaments
to accommodate logistical differences.

\subsection{In-Person Tournament}

The in-person tournament used a full serpentine (``snake'') draft before each
round. With fewer teams than available slots for \ai{} agents (two per team),
each team selected two \ai{} teammates without conflict.
The serpentine ordering ensured competitive balance:
the lowest-scoring team picked first, selection proceeded upward to
the highest-scoring team (who picked twice consecutively), then reversed back
down. This gave weaker teams first access to perceived stronger \ai{} agents.

\subsection{Online Tournament}

The online tournament had more team--game slots than distinct \ai{} systems.
To prevent the same \ai{} agent from facing itself in a match, we modified the
draft to operate per-game rather than per-round. Before each game, the two
competing teams drafted their \ai{} teammates from the available pool, with
agents selected by one team becoming unavailable to the opponent for that game.
This ensured no \ai{} agent appeared on both sides of the same match.

\subsection{Playoff and Finals Modifications}

During playoff rounds in both tournaments, we adopted the same per-game draft
style as the online tournament. This allowed teams participating in finals and
playoffs to have their best shot at selecting optimal \ai{} teammates while
adhering to the constraint that no two \ai{} systems could face each other in
the same game.

\subsection{Implications for Analysis}

These draft variations do not affect our main analyses, which focus on
within-game reliance decisions (muting, switching) rather than cross-game
selection patterns. However, the draft mechanism itself provides an
\draftSelection{} signal: teams' choices reveal their beliefs about \ai{}
capabilities based on observed performance in earlier rounds. We leave detailed
analysis of draft strategy to future work.

\section{Explanation Feature Analysis}
\label{appendix:collabqa-explanation-features}

This appendix details the feature extraction and analysis for \ai{}
explanations presented in Section~\ref{subsec:collabqa-bonus-analysis}. We extract
57 features from each explanation (49 statistical + 8 LLM-assessed) and
evaluate their predictive power for two outcomes: (1) whether the \ai{}
answer is correct, and (2) whether humans select that explanation.

\subsection{Experimental Setup}

For each feature, we fit a single-feature logistic regression and measure
prediction accuracy. This isolates each feature's
individual contribution, avoiding confounds from correlated predictors.

\paragraph{\ai{} Correctness Prediction.}
We use all bonus responses where both \ai{} systems provided answers. The
outcome is binary: whether the \ai{}'s answer matches the gold answer.

\paragraph{Human Selection Prediction.}
We analyze cases where humans chose between two \ai{} explanations. The
outcome is which explanation was selected.

\subsection{Feature Categories}

We organize features into seven categories with consistent prefixes for
filtering and analysis.

\paragraph{LLM-Assessed Features.}
These eight features are rated by GPT-4o on a 1--5 scale, normalized to 0--1.
Among these, \texttt{question\_comprehension} is the top correctness predictor
at 76\%, while \texttt{evidential\_grounding} is the only feature appearing
in top-10 for both correctness and human selection. The \texttt{overconfidence}
feature serves as a negative predictor of correctness.

\begin{table}[h]
\centering
\scriptsize
\begin{tabular}{p{3.2cm}p{4cm}}
\toprule
\textbf{Feature} & \textbf{Description} \\
\midrule
\texttt{question\_comprehension} & Understanding of what question asks \\
\texttt{evidential\_grounding} & Cites specific clues as evidence \\
\texttt{reasoning\_coherence} & Logical flow from clues to answer \\
\texttt{domain\_expertise\_display} & Domain-specific knowledge shown \\
\texttt{reasoning\_transparency} & Reasoning process explicit/followable \\
\texttt{reasoning\_depth} & Multi-step vs.\ surface reasoning \\
\texttt{answer\_explanation\_alignment} & Explanation supports stated answer \\
\texttt{overconfidence} & Unwarranted certainty given evidence \\
\bottomrule
\end{tabular}
\caption{LLM-assessed features (\texttt{llm\_*}).}
\label{tab:collabqa-features-llm}
\end{table}

\paragraph{Epistemic Features.}
These six features capture uncertainty expression. The
\texttt{confidence\_alignment} feature---measuring match between linguistic
and numeric confidence---is a strong correctness predictor.

\begin{table}[h]
\centering
\scriptsize
\begin{tabular}{p{3.2cm}p{4cm}}
\toprule
\textbf{Feature} & \textbf{Description} \\
\midrule
\texttt{confidence\_alignment} & Match: linguistic vs.\ numeric confidence \\
\texttt{hedge\_ratio} & Ratio of hedging words (might, could, perhaps) \\
\texttt{certainty\_ratio} & Ratio of certainty markers (definitely, clearly) \\
\texttt{uncertainty\_ratio} & Ratio of uncertainty markers (unsure, unclear) \\
\texttt{linguistic\_confidence} & Net confidence from language \\
\texttt{modal\_verb\_count} & Count of modal verbs \\
\bottomrule
\end{tabular}
\caption{Epistemic features (\texttt{epist\_*}).}
\label{tab:collabqa-features-epistemic}
\end{table}

\paragraph{Content Grounding Features.}
These eight features measure how explanations relate to question content.
Notably, humans rely heavily on \texttt{semantic\_similarity} (66\%) and
\texttt{word\_overlap\_ratio} (63\%) despite these being weak correctness
signals---a key driver of the calibration gap.

\begin{table}[h]
\centering
\scriptsize
\begin{tabular}{p{3.2cm}p{4cm}}
\toprule
\textbf{Feature} & \textbf{Description} \\
\midrule
\texttt{semantic\_similarity} & BERT embedding cosine similarity \\
\texttt{jaccard\_similarity} & Jaccard similarity of word sets \\
\texttt{word\_overlap\_ratio} & Shared words / explanation words \\
\texttt{word\_overlap\_count} & Raw count of shared words \\
\texttt{shared\_entities} & Named entities in both texts \\
\texttt{exp\_entity\_count} & Total entities in explanation \\
\texttt{exp\_entity\_density} & Entity count / token count \\
\texttt{mentions\_answer\_text} & Answer string appears in explanation \\
\bottomrule
\end{tabular}
\caption{Content grounding features (\texttt{content\_*}).}
\label{tab:collabqa-features-content}
\end{table}

\paragraph{Surface Linguistic Features.}
These 15 features capture basic textual properties. Length measures show
that longer explanations are not necessarily better.

\begin{table}[h]
\centering
\scriptsize
\begin{tabular}{p{3.2cm}p{4cm}}
\toprule
\textbf{Feature} & \textbf{Description} \\
\midrule
\texttt{word\_count} & Total word count \\
\texttt{sentence\_count} & Total sentence count \\
\texttt{avg\_word\_length} & Average characters per word \\
\texttt{avg\_sentence\_length} & Average words per sentence \\
\texttt{type\_token\_ratio} & Lexical diversity \\
\texttt{flesch\_reading\_ease} & Readability (higher = easier) \\
\texttt{flesch\_kincaid\_grade} & Grade level required \\
\texttt{smog\_index} & SMOG readability index \\
\texttt{ari} & Automated readability index \\
\texttt{noun\_ratio} & Proportion of nouns \\
\texttt{verb\_ratio} & Proportion of verbs \\
\texttt{adj\_ratio} & Proportion of adjectives \\
\texttt{adv\_ratio} & Proportion of adverbs \\
\bottomrule
\end{tabular}
\caption{Surface linguistic features (\texttt{surface\_*}).}
\label{tab:collabqa-features-surface}
\end{table}

\paragraph{Structural Features.}
These nine features capture formatting and discourse structure. The
\texttt{has\_quotes} feature is the top human predictor at 70\%, yet
provides a weak correctness signal---another driver of miscalibration.

\begin{table}[h]
\centering
\scriptsize
\begin{tabular}{p{3.2cm}p{4cm}}
\toprule
\textbf{Feature} & \textbf{Description} \\
\midrule
\texttt{has\_quotes} & Contains quotation marks \\
\texttt{clue\_mentions} & References to ``clue,'' ``line,'' ``phrase'' \\
\texttt{position\_mentions} & References to ``first,'' ``beginning,'' etc. \\
\texttt{mentions\_answer} & Contains ``answer,'' ``guess,'' ``conclude'' \\
\texttt{num\_clauses} & Syntactic complexity \\
\texttt{causal\_connective\_ratio} & Because, therefore, thus, hence \\
\texttt{contrastive\_connective\_ratio} & However, but, although, yet \\
\texttt{additive\_connective\_ratio} & And, also, moreover, furthermore \\
\texttt{has\_parentheses} & Contains parenthetical asides \\
\bottomrule
\end{tabular}
\caption{Structural features (\texttt{struct\_*}).}
\label{tab:collabqa-features-structural}
\end{table}

\paragraph{Reasoning Pattern Features.}
These five features detect specific reasoning strategies through lexical
patterns.

\begin{table}[h]
\centering
\scriptsize
\begin{tabular}{p{3.2cm}p{4cm}}
\toprule
\textbf{Feature} & \textbf{Description} \\
\midrule
\texttt{elimination} & ``rules out,'' ``can't be,'' ``eliminate'' \\
\texttt{pattern\_matching} & ``characteristic of,'' ``typical,'' ``consistent with'' \\
\texttt{analogical} & ``similar to,'' ``like,'' ``resembles'' \\
\texttt{definitional} & ``defined as,'' ``means,'' ``refers to'' \\
\texttt{abductive} & ``best explains,'' ``most likely,'' ``suggests'' \\
\bottomrule
\end{tabular}
\caption{Reasoning pattern features (\texttt{reason\_*}).}
\label{tab:collabqa-features-reasoning}
\end{table}

\paragraph{Pragmatic Features.}
These six features capture interaction style and metacognition.

\begin{table}[h]
\centering
\scriptsize
\begin{tabular}{p{3.2cm}p{4cm}}
\toprule
\textbf{Feature} & \textbf{Description} \\
\midrule
\texttt{metacognitive\_markers} & ``I think,'' ``I believe,'' ``uncertain'' \\
\texttt{addressee\_references} & Direct ``you'' references \\
\texttt{imperative\_count} & ``consider,'' ``note,'' ``look,'' ``see'' \\
\texttt{alternative\_mentions} & ``also,'' ``alternatively,'' ``could be'' \\
\texttt{conditional\_count} & ``if'' statements \\
\texttt{limitation\_admission} & ``don't know,'' ``not sure,'' ``might be wrong'' \\
\bottomrule
\end{tabular}
\caption{Pragmatic features (\texttt{pragma\_*}).}
\label{tab:collabqa-features-pragmatic}
\end{table}

\subsection{Full Feature Rankings}

\begin{figure*}[t]
    \centering
    \includegraphics[width=\textwidth]{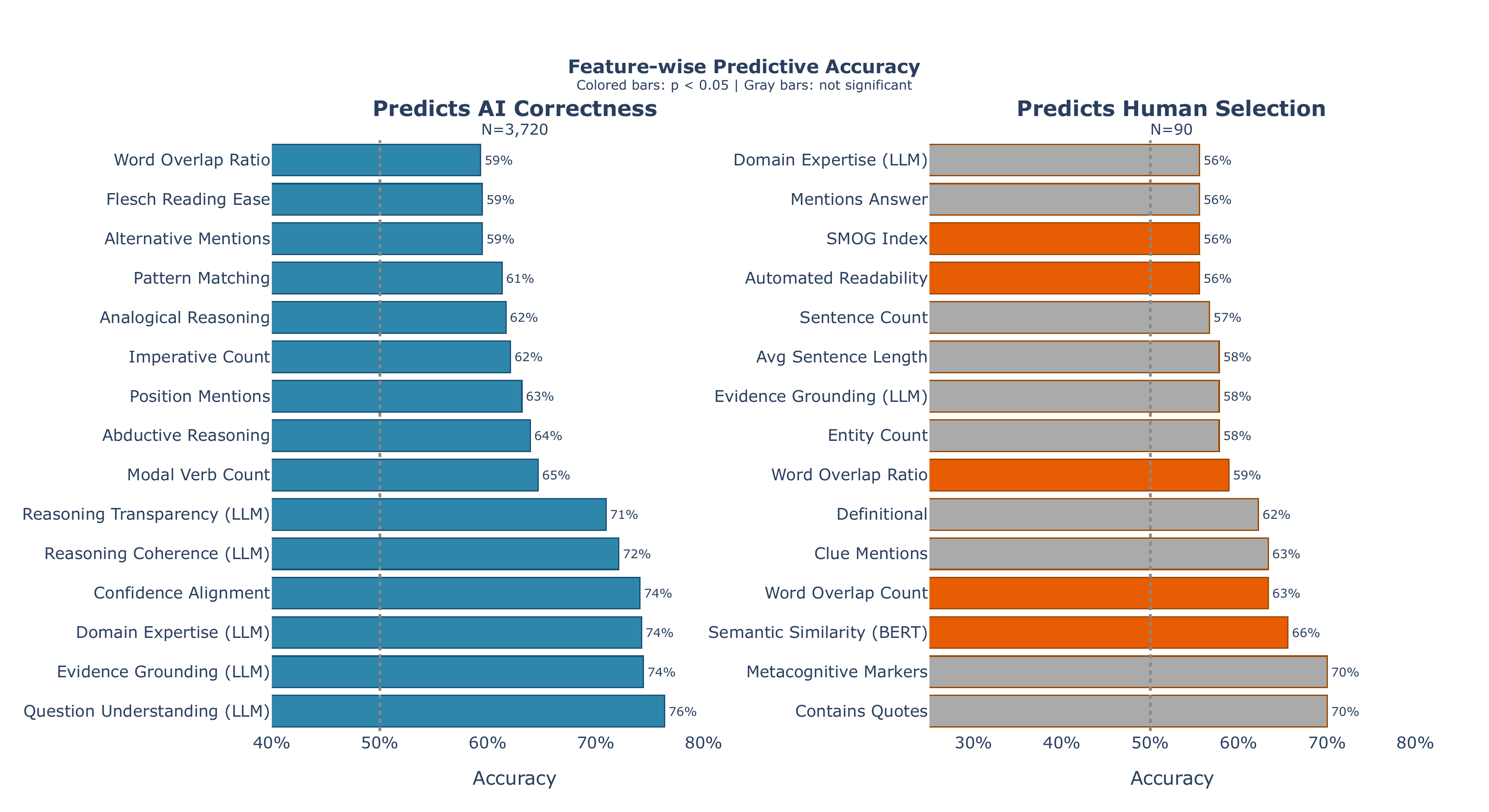}
    \caption{Full feature-wise prediction accuracy. Left: predicting \ai{}
    correctness. Right: predicting human selection. Colored bars indicate
    statistical significance ($p < 0.05$); gray bars are not significant.
    LLM-assessed features dominate correctness prediction but are largely
    ignored by humans.}
    \label{fig:explanation-features-full}
\end{figure*}

Figure~\ref{fig:explanation-features-full} shows single-feature prediction
accuracy for all features across both outcomes. LLM-assessed features
cluster at high correctness accuracy (70--76\%) but near-chance human
prediction (44--58\%). Surface and structural features show the opposite
pattern.

\subsection{Key Takeaways}

\begin{enumerate}
    \item \textbf{One shared signal:} Only \texttt{evidential\_grounding}
    appears in both top-10 lists, suggesting humans recognize quality when
    reasoning explicitly cites evidence.
    
    \item \textbf{LLM features predict correctness:}
    \texttt{question\_comprehension} (76\%), \texttt{evidential\_grounding}
    (74\%), \texttt{domain\_expertise\_display} (74\%), and
    \texttt{reasoning\_coherence} (72\%) strongly predict whether the \ai{}
    is correct.
    
    \item \textbf{Surface features predict human trust:}
    \texttt{has\_quotes} (70\%), \texttt{semantic\_similarity} (66\%), and
    \texttt{word\_overlap\_ratio} (63\%) predict human selection but are
    weak correctness signals.
    
    \item \textbf{Implication:} \ai{} explanations should make reasoning
    explicit through evidence citation. Humans should evaluate reasoning
    quality rather than surface familiarity.
\end{enumerate}

\subsection{Implementation Details}

Features are extracted using a modular pipeline with seven extractors:

\begin{itemize}
    \item \textbf{SurfaceLinguisticExtractor}: Uses spaCy for tokenization
    and POS tagging; textstat for readability indices.
    
    \item \textbf{StructuralExtractor}: Pattern matching for discourse
    connectives and formatting signals.
    
    \item \textbf{ContentGroundedExtractor}: Uses sentence-transformers
    (all-MiniLM-L6-v2) for semantic similarity; spaCy for entity extraction.
    
    \item \textbf{ReasoningTypeExtractor}: Regex patterns for reasoning
    strategy indicators.
    
    \item \textbf{EpistemicExtractor}: Lexicon-based detection of hedges,
    certainty markers, and modal verbs.
    
    \item \textbf{PragmaticExtractor}: Pattern matching for metacognitive
    and interaction markers.
    
    \item \textbf{LLMBasedExtractor}: GPT-4o with structured prompting for
    semantic quality assessment (1--5 scale, normalized to 0--1).
\end{itemize}

All features use consistent prefixes (\texttt{surface\_}, \texttt{struct\_},
\texttt{content\_}, \texttt{reason\_}, \texttt{epist\_}, \texttt{pragma\_},
\texttt{llm\_}) enabling category-based filtering and analysis.

\section{\ai{} Assistant Systems}
\label{appendix:collabqa-ai-systems}

This appendix details the \ai{} assistant systems used in the tournament.
Each \ai{} teammate consists of two sub-agents: one for the \tossup{} (tossup)
phase and one for the \bonus{} (bonus) phase, reflecting the distinct
requirements of each question mode.

\subsection{System Collection}

We collected \ai{} systems through a four-week online competition prior to
tournament play.
Participants submitted their systems to our evaluation platform, where they
were tested on a held-out set of 80 questions per mode (separate from the
questions used in the live tournament).
This competition format encouraged diverse architectural approaches while
providing participants time to iterate on their designs.
The resulting systems span a range of strategies, from single-model
configurations with carefully tuned prompts to multi-step pipelines involving
text-analysis, ensemble voting, and confidence calibration.
During the tournament, teams draft these \ai{} teammates across different
rounds using the serpentine selection process described in
Section~\ref{sec:collabqa-tournament}.

\subsection{Tossup Agent Specifications}

In the \tossup{} phase, agents receive a partial prefix of the question as it
is read aloud, since it is supposed to be \emph{interrupted}.
At each word boundary, the agent outputs two values: a boolean \emph{buzz}
decision and its most confident \emph{guess} given the clues seen so far.
Our moderator stops reading the question when the model first buzzes and accepts the guess at
that point as the team's response.
Table~\ref{tab:collabqa-tossup-agents} details each system's architecture for this
phase, where agents must decide when to buzz and provide answers under time
pressure.
The systems exhibit considerable variation in their approach to confidence
calibration and answer generation.

\begin{table*}[t]
\centering
\small
\begin{tabularx}{\textwidth}{l>{\raggedright\arraybackslash}p{4.2cm}cX}
\toprule
\textbf{System} & \textbf{Models Used} & \textbf{Calls} & \textbf{Approach} \\
\midrule
\rowcolor{gray!10}
\RodeRunner{} &
GPT-4o (answer) &
1 &
\emph{Single-shot} solver using prompt-level calibration norms
and probability gating for latency optimization. \\
\addlinespace[3pt]
\BlackRaven{} &
GPT-4.1 (answer) &
1 &
\emph{Rule-intensive} Quizbowl specialist with domain norms embedded in prompts,
including indicator discipline and discrete confidence scales. \\
\addlinespace[3pt]
\rowcolor{gray!10}
\Tigerclaw{} &
GPT-4.1-nano (word count),
GPT-4.1-nano (entity count),
GPT-4.1 (answer),
GPT-4.1-mini (calibration) &
4 &
\emph{Confidence engineering} pipeline that decouples correctness from
certainty, rescaling confidence using question completeness heuristics. \\
\addlinespace[3pt]
\WiseWings{} &
GPT-4o (voter),
GPT-3.5-turbo (voter),
GPT-4o (aggregator) &
3 &
\emph{Exact-match voting} across heterogeneous models; abstains
with zero confidence unless all voters fully agree. \\
\addlinespace[3pt]
\rowcolor{gray!10}
\Sphinx{} &
Claude-3.5-Sonnet (analysis),
GPT-4o (answer),
Command-R-Plus (confidence) &
3 &
\emph{Structured pipeline} separating clue analysis,
answer generation, and confidence assignment. \\
\addlinespace[3pt]
\Sam{} &
Command-R (extraction),
GPT-4.1-mini (answer),
Claude-3.5-Haiku (verifier),
GPT-4o-mini (aggregator) &
4 &
\emph{Verifier-centered} architecture treating generation as hypothesis,
penalizing disagreement via AND-style aggregation. \\
\addlinespace[3pt]
\rowcolor{gray!10}
\Magicarp{} &
GPT-4o-mini (candidate),
Claude-3.5-Haiku (candidate),
GPT-4.1-mini (cross-check),
Command-R (confidence) &
4 &
\emph{Cross-validation} combining independently proposed answers with
match-based confidence scoring. \\
\addlinespace[3pt]
\Bayleef{} &
Command-R (answer) &
1 &
\emph{Single-pass} Cohere-native design emphasizing disciplined uncertainty
reporting and probability thresholds. \\
\bottomrule
\end{tabularx}
\caption{Tossup (proactive phase) agent specifications. Each system
implements a distinct strategy for answer generation and confidence
calibration under the time pressure of buzzer-style questions.}
\label{tab:collabqa-tossup-agents}
\end{table*}

\subsection{Bonus Agent Specifications}

In the \bonus{} phase, agents receive the lead-in context and the current
part of a three-part question.
For each part, the agent outputs three values: a \emph{guess}, a
\emph{confidence} score, and an \emph{explanation} justifying the answer.
Table~\ref{tab:collabqa-bonus-agents} details each system's architecture for this
phase.
These configurations often differ from their \tossup{} counterparts,
reflecting the different demands of deliberative question answering.

\begin{table*}[t]
\centering
\small
\begin{tabularx}{\textwidth}{l>{\raggedright\arraybackslash}p{4.2cm}cX}
\toprule
\textbf{System} & \textbf{Models Used} & \textbf{Calls} & \textbf{Approach} \\
\midrule
\rowcolor{gray!10}
\RodeRunner{} &
GPT-4o-mini (answer) &
1 &
\emph{Single-shot} lightweight model with strict JSON-only output format. \\
\addlinespace[3pt]
\BlackRaven{} &
GPT-4o (answer) &
1 &
\emph{High-accuracy} model with Quizbowl-specific guardrails for answer formatting. \\
\addlinespace[3pt]
\rowcolor{gray!10}
\Tigerclaw{} &
DeepSeek V3 (answer) &
1 &
\emph{End-to-end} single DeepSeek model for all outputs. \\
\addlinespace[3pt]
\WiseWings{} &
GPT-4o (analyzer),
GPT-4o (generator),
GPT-4o (evaluator) &
3 &
\emph{Multi-role} pipeline reusing same model across analyzer, generator, and evaluator. \\
\addlinespace[3pt]
\rowcolor{gray!10}
\Sphinx{} &
GPT-4o-mini (advisor) &
1 &
\emph{Advisory-style} solver providing answers with explanations. \\
\addlinespace[3pt]
\Sam{} &
Command-R (extraction),
GPT-4.1-mini (generation),
GPT-4o-mini (verification),
Claude-3.5-Haiku (aggregation) &
4 &
\emph{Multi-model pipeline} with distinct roles: extraction, generation,
verification, and aggregation. \\
\addlinespace[3pt]
\rowcolor{gray!10}
\Magicarp{} &
Claude-3.5-Haiku (hypothesis),
GPT-4o-mini (scoring),
GPT-4.1-mini (calibration),
Command-R (explanation) &
4 &
\emph{Sequential synthesis} for hypothesis generation, scoring,
calibration, and explanation. \\
\addlinespace[3pt]
\Bayleef{} &
Command-R (answer) &
1 &
\emph{Single-pass} Cohere model with strict JSON-only output constraints. \\
\bottomrule
\end{tabularx}
\caption{Bonus (deliberative phase) agent specifications. Systems vary from
single-model configurations to multi-step pipelines, reflecting different
strategies for collaborative question answering with explanations.}
\label{tab:collabqa-bonus-agents}
\end{table*}

\section{Tournament Platform and Data Release}
\label{appendix:collabqa-software}

\subsection{Tournament Web Application}

Both tournaments were conducted using a custom web application
built for human--\ai{} collaborative quizbowl. The platform
provides:

\begin{itemize}
\item \textbf{Real-time question display} with
  clue-by-clue progression for tossup questions, mirroring
  the pacing of a live moderator.

\item \textbf{Buzzer integration} via BuzzIn Live for the
  online tournament (physical buzzers for the in-person
  event), supporting simultaneous human and \ai{} buzz
  attempts with millisecond-resolution timestamps.

\item \textbf{Bonus collaboration interface}
  (Figure~\ref{fig:bonus-interface}) presenting human
  guess entry, \ai{} suggestions with confidence and
  explanations, and final answer submission in a
  three-step flow.

\item \textbf{Moderator dashboard} for controlling game
  flow, recording answer correctness, managing the
  serpentine draft, and tracking scores in real time.
\end{itemize}

The online tournament was conducted over a Zoom call with
the moderator sharing the web application screen and
managing game progression through the dashboard. All
participant interactions with the interface were logged
automatically.

\subsection{Model Submission Platform}

We received approximately 60 tossup agent submissions and
25 bonus agent submissions through a four-week open
competition. Agents were evaluated on a held-out set of 80
questions from a recent tournament-difficulty quizbowl
question set. Tossup agents were ranked by expected points
against human buzz-point logs; bonus agents by average
part-level accuracy. We paired each submitter's top tossup
and bonus agents into a single ``quizbowl \ai{} agent''
and selected the top 8 paired agents, ensuring
architectural diversity across base models and prompting
strategies. See Appendix~\ref{appendix:collabqa-ai-systems}
for per-agent specifications.

\subsection{Data and Code Release}

We release:

\begin{enumerate}
\item \textbf{Dataset:} All tournament questions
  (tossup and bonus), human responses at each stage
  (initial guess, final answer), \ai{} responses
  (answers, confidence scores, explanations), muting
  decisions, answer correctness labels, and anonymized
  player metadata (experience level, team assignment).
  All personal and identifiable information has been
  explicitly anonymized before release. Available at
  \url{\datasetUrl{}}.

\item \textbf{Tournament application:} Source code for
  the web application, including the buzzer integration,
  bonus collaboration interface, moderator dashboard, and
  draft management system. Available at \url{\systemUrl{}}.

\item \textbf{Analysis scripts:} All scripts used for
  feature extraction, statistical testing, figure
  generation, and the logistic regression analyses
  reported in this paper. Available at \url{\githubUrl{}}.
\end{enumerate}

\end{document}